\documentclass{article}

\pdfoutput=1

\usepackage{PRIMEarxiv}

\usepackage[utf8]{inputenc} 
\usepackage[T1]{fontenc}    
\usepackage{hyperref}       
\usepackage{url}            
\usepackage{booktabs}       
\usepackage{amsmath,amsfonts}       
\usepackage{nicefrac}       
\usepackage{microtype}      
\usepackage{lipsum}
\usepackage{fancyhdr}       
\usepackage{graphicx}       
\usepackage{mathtools} 
\usepackage{booktabs} 
\usepackage{tikz} 
\usepackage{algpseudocode, algorithm}
\usepackage{wrapfig}
 \usepackage[stable]{footmisc}
 
\newcommand {\myvec}[1] {{\mbox{\boldmath $#1$}}}
\newcommand{\myth}{\myvec{\theta}}
\newcommand{\mys}{\myvec{s}}
\newcommand{\prob}{\mathbb{P}}

\usepackage{lipsum}  
\usepackage[normalem]{ulem}
\newcommand{\ul}[1]{\uline{#1}}

\newcommand{\textrmtt}{\textnormal\texttt}

\usepackage{amsfonts,amsmath,bm,amsthm}

\makeatletter
\renewcommand{\ALG@beginalgorithmic}{\small}
\makeatother

\usepackage{natbib} 
\usepackage{microtype}
\usepackage{graphicx}
\usepackage{subfigure}
\usepackage{booktabs} 

\usepackage{hyperref}
\usepackage{amsthm}
\theoremstyle{plain}

\theoremstyle{definition}

\theoremstyle{remark}

\newtheoremstyle{boldtheorem} 
  {3pt} 
  {2pt} 
  {} 
  {} 
  {\bfseries} 
  {.} 
  {.5em} 
  {} 

\theoremstyle{boldtheorem}

\usepackage[textsize=tiny]{todonotes}
\usepackage{bm}
\usepackage{derivative}
\usepackage{amsmath}
\usepackage{amsthm}
\usepackage{amssymb}
\usepackage{mathtools}

\usepackage{multirow}
\usepackage{multicol}
\usepackage{afterpage}
\usepackage{graphicx}
\usepackage{thmtools,thm-restate}
  
\title{Contextual Feature Selection with Conditional Stochastic Gates
}

\author{
  Ram Dyuthi Sristi \\
  UC San Diego \\
  La Jolla, CA, USA\\
  \textrmtt{rsristi@ucsd.edu} \\
   \And
  Ofir Lindenbaum \\
  Bar-Ilan University \\
  Ramat Gan, Israel\\
  \textrmtt{ofirlin@gmail.com} \\
  \And
  Shira Lifshitz \\
  Technion \\
  Haifa, Israel \\
   \textrmtt{shiralif@campus.technion.ac.il} \\
 \And
  Maria Lavzin\\
  Technion \\
  Haifa, Israel \\
  \textrmtt{maria.lavzin@gmail.com} \\
   \And
   Jackie Schiller\\
  Technion \\
  Haifa, Israel \\
 \textrmtt{jackie@technion.ac.il} \\
  \And
  Gal Mishne \\
  UC San Diego \\
  La Jolla, CA, USA\\
  \textrmtt{gmishne@ucsd.edu} \\
  \And
  Hadas Benisty \\
  Technion \\
  Haifa, Israel \\
  \textrmtt{hadasbe@technion.ac.il } \\
}

\begin{document}
\maketitle

\begin{abstract}
\looseness = -1
Feature selection is a crucial tool in machine learning and is widely applied across various scientific disciplines. Traditional supervised methods generally identify a universal set of informative features for the entire population. However, feature relevance often varies with context, while the context itself may not directly affect the outcome variable. Here, we propose a novel architecture for contextual feature selection where the subset of selected features is conditioned on the value of \textit{context variables}. Our new approach, Conditional Stochastic Gates (c-STG), models the importance of features using conditional Bernoulli variables whose parameters are predicted based on contextual variables. We introduce a hypernetwork that maps context variables to feature selection parameters to learn the context-dependent gates along with a prediction model. We further present a theoretical analysis of our model, indicating that it can improve performance and flexibility over population-level methods in complex feature selection settings. Finally, we conduct an extensive benchmark using simulated and real-world datasets across multiple domains demonstrating that c-STG can lead to improved feature selection capabilities while enhancing prediction accuracy and interpretability.
\end{abstract}

\section{Introduction}

Feature selection techniques are vital in Machine Learning (ML) as they identify informative features from large sets of observed variables. These techniques are increasingly crucial across scientific domains due to the high dimensionality of collected data and complex prediction models. Feature selection simplifies models by removing nuisance features and identifying informative features, ultimately improving generalization \cite{Li_2017, Kumar2014FeatureSA, islam2022comprehensive, everaert2022features}. Feature selection can be broadly categorized into filter, wrapper, and embedded methods.

Filter methods~\cite{MI1,MI2,MI3,HSIC1,HSIC2,HSIC3,sristidisc} use a predefined criterion, independent of the predictive model, to rank and select features mostly based on statistical measures such as correlation or mutual information~\cite{lewis1992feature}. 
Wrapper methods~\cite{wrapper1,wrapper2,wrapper3,wrapper4,KernelW}, on the other hand, use the predictive model itself as a criterion for feature selection. These methods select subsets of features and evaluate the prediction quality based only on those features to identify the subset with the best performance. This can be prohibitively expensive for complex models. 
Embedded methods~\cite{Lasso,Lasso2,Lasso3,Lasso4} incorporate feature selection into the training process of the predictive model, for example, regularization-based techniques like LASSO and its variants \cite{daubechies2008iteratively,bertsimas2017trimmed}. Here we present a novel embedded method to learn the prediction model and informative features in an end-to-end fashion.

\begin{figure*}[t]
\centering
\includegraphics[width=\textwidth]{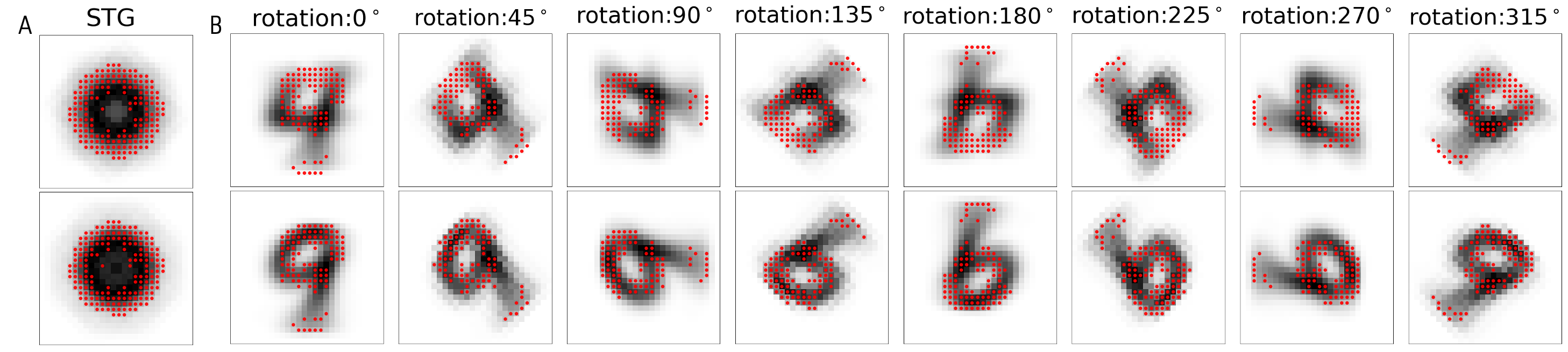} 
    \caption{Illustration with ``rotating MNIST''. We perform a binary classification between rotated versions of the digits 4 (top row) and 9 (bottom row) from the MNIST dataset. 
    We compare the features selected by the global STG \cite{yamada2020feature} model (A) and our proposed c-STG (B), which selects features conditioned on the rotation angle.
    Each image depicts the mean pixel values across all rotated images (A) or all images at a given rotation angle (B). Red dots indicate the features selected using STG (A) or c-STG (B) for each rotation angle. c-STG can learn to dynamically change its prediction of the most informative features given the context (rotation angle).}
    \label{fig:mnist}

\end{figure*}

The methods above are global in the sense that they identify a single set of informative features for the entire data set. However, in certain cases, the importance of features varies depending on contextual information, while the context itself does not encode the outcome variable on its own. 
For example, in healthcare, feature selection reveals which explanatory features directly predict disease risk, but feature relevance and importance can vary based on context, such as age or gender. Thus, considering context adds depth to medical analysis and offers clinicians valuable insights. Similarly, product recommendation systems need to tailor feature selection to the user's location, time, and device~\cite{baltrunas2009context}.

Many feature selection methods fail to fully address the intricate relationship between features, context, and outcomes, often neglecting contextual variables or merely concatenating them with explanatory features. This approach hinders interpretability, as it does not clearly show the dependency of features on the context. An alternative is to train a separate model for different values of categorical contextual variables, e.g., gender, by dividing the data accordingly. However, this strategy significantly increases computational demands with multiple categorical contexts and decreases the available training data for each model, potentially affecting performance. Furthermore, this method necessitates arbitrary binning for continuous contextual variables (e.g., age, location), which coarsens the interpretability of feature importance across contexts. This issue becomes more pronounced when dealing with multiple contextual variables, making it challenging to maintain nuanced interpretability.

For illustration purposes, we consider a binary classification task between digits 4 and 9 observed at various rotations (see Fig. \ref{fig:mnist}).
Suppose the goal is to select a subset of informative pixels (features) for classifying the two digits. 
In the original orientation (zero rotation), the pixels on the top of the image are more significant than those at the bottom for distinguishing between 4 and 9. 
When we rotate the digits by intervals of 45 degrees, global feature selection models will select almost all features as informative. 
However, given a specific rotation angle, only a subset of these features are informative. Moreover, the rotation angle is unrelated to the response (the identity of the digit). Using rotation as a contextual variable to select the informative pixels for the classification can alleviate overfitting and improve the interpretability of the model.
In Figure~\ref{fig:mnist} (B), we present features (pixels) selected by our proposed approach, conditioned on a rotation variable. The significant features rotate as the context changes, and our model identifies the features that best distinguish between digits 4 and 9 for each context. In contrast, the features selected by the global STG~\cite{yamada2020feature} (Fig.~\ref{fig:mnist} A), trained across all rotated images, mainly recover the action of rotation, without providing meaningful interpretable features for class separation.   

\begin{figure*}[t]
    \centering
    \includegraphics[width=0.86\textwidth]{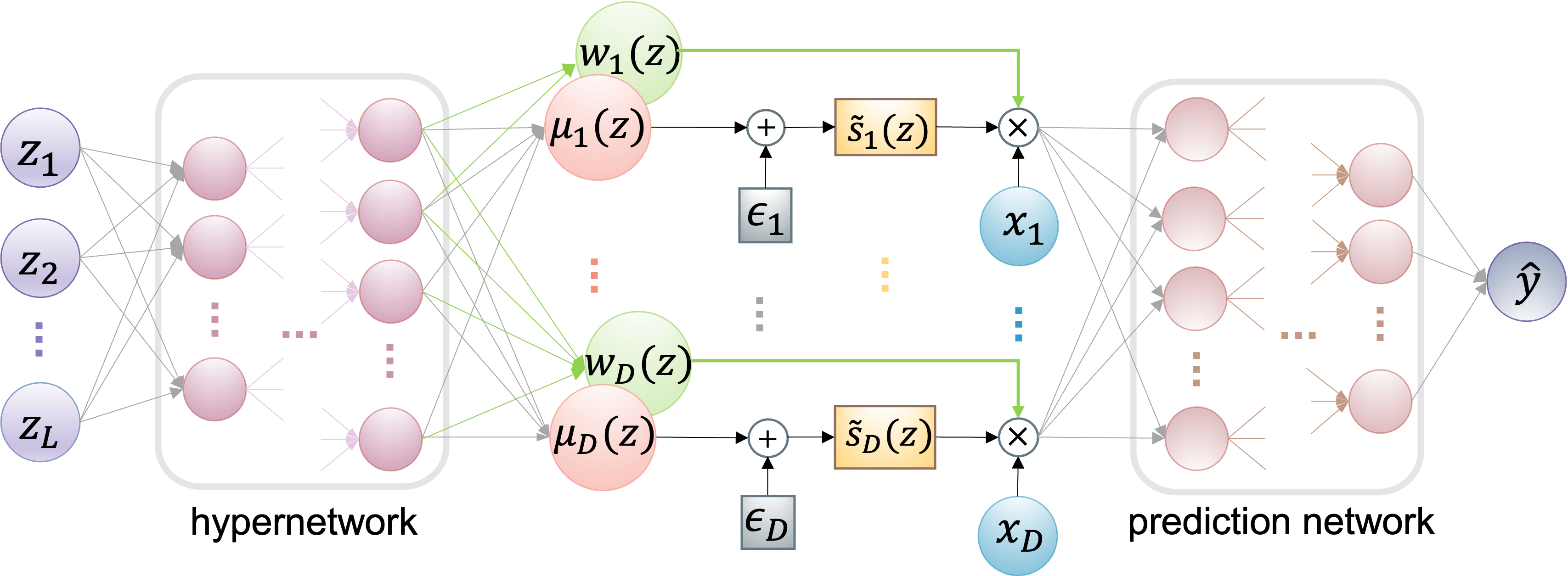}
    \caption{Contextual feature selection framework. Contextual variables $z$ (in purple) feed into the hypernetwork. The hypernetwork outputs the parameters of the gates, $\mu(z)$, which are combined with $\epsilon$ to determine if each gate is open or close $\tilde{s}_d$ (yellow) for each feature $x_d$ (blue). For weighted c-STG, the hypernetwork also outputs weight vectors (green), indicating the importance of the selected explanatory features. The selected and weighted features are fed into the prediction model, thus enhancing its ability to process feature significance in predictions. 
    }
    \label{fig:blockdiagram}
\end{figure*}

In this paper, we propose c-STG, a method for context-specific feature selection in a supervised learning setup. 
We posit that the informativeness of a feature is governed by a conditional Bernoulli distribution, aiming to learn its parameters to enhance feature selection. To enable parameter learning via backpropagation, we reparametrize the discrete Bernoulli variables using a truncated Gaussian~\cite{yamada2020feature} whose mean is predicted using a hypernetwork that receives the context variables as input. 
Furthermore, we propose \textit{weighted} c-STG, which uses a hypernetwork to learn both a score for the significance of the selected features and the parameters of the probabilistic feature selection. In both c-STG and weighted c-STG, we learn the weights of the hypernetwork and the weights of a prediction model by empirical risk minimization.
Since the hypernetwork is parametric, we can generalize the contextual feature selection to unseen contexts. 
Our framework enables studying the explainability of the model along two axes: 1) given a context, we identify the importance of the different features, which is essential for performing an accurate prediction task; and 2) given a feature, we identify how its importance varies across different contexts as well as interpolate or extrapolate to unseen contexts.  
Overall, our contributions are as follows:
\vskip -0.06 in
\noindent 1. We develop an end-to-end machine learning framework comprising two distinct but interconnected networks: a hypernetwork and a prediction network. In our c-STG, the hypernetwork maps contextual variables to feature selection parameters, and the prediction network links these selected features to the outcome. c-STG can handle categorical, continuous, and/or multi-dimensional contextual variables.
\vskip -0.06 in
\noindent 2.  In weighted c-STG, we augment the c-STG hypernetwork to further output a context-dependent weight vector, evaluating each feature's impact. Thus, the prediction network leverages the selected and weighted features, improving the prediction by combining feature selection and significance.
\vskip -0.06 in
\noindent 3. We analyze the optimal risk of c-STG compared with STG and study the optimal solutions under a linear regression prediction model. 
\vskip -0.06 in
\noindent 4. We conduct comprehensive empirical evaluations on simulated and real-world datasets across healthcare, housing, and neuroscience, demonstrating the effectiveness and adaptability of our proposed methods compared to existing techniques. 

\noindent \textbf{Related Work:}

Several solutions have been proposed for the problem of context-specific feature selection, including Contextual Explanation Networks (CEN)~\cite{alshedivat2020contextual}, Contextualized ML~\cite{lengerich2023contextualized}, and Contextual LASSO~\cite{thompson2023contextual}. These methods usually consider the prediction task as a linear function of explanatory features and determine the model's parameters based on contextual variables. 
In contrast, our method accommodates both linear and non-linear models, thus extending the applicability beyond the constraints of traditional methods.
While CEN and Contextualized ML do not employ sparsity constraints, Contextual LASSO utilizes an $\ell_1$ regularization that leads to shrinkage of the model's coefficients. In contrast, our c-STG approach uses stochastic gates regularization, which effectively approximates the $\ell_0$ sparsity constraint in a context-conditioned manner, therefore achieving sparser solutions.

Recently, several dynamic feature selection approaches have been proposed.
In sample-wise methods~\cite{chen2018learning,yoon2018invase,pmlr-v162-yang22i} 
feature selection is tailored to each sample based on its feature values, requiring all feature values to be known in advance. This approach could be limiting in areas like healthcare, where not all possible test results are available for the entire population. 
Active learning-based methods~\cite{shim2018joint, covert2023learning} focus on real-time feature acquisition and iteratively select features for each sample. These methods, though not originally designed for contextual feature selection, can incorporate context by concatenating contextual and explanatory features. Both sample-wise and active learning methods can be used in context-dependent setups. In sample-specific selection, one can ensure contextual variables are automatically deemed significant from the start, subsequently determining the significance of the remaining explanatory features. 
In active selection methods, contextual variables can be prioritized before choosing significant explanatory features, ensuring a context-informed selection process. Both approaches empower the model to identify essential features within the specific context, ensuring context-driven, tailored analysis per sample.

The interpretability of these models, however, is not straightforward. Identifying which features are selected as a function of context would require aggregating the selected features across all samples per context value, thus resorting to coarse binning. 
In contrast, our c-STG discerns crucial features based on context and can adapt to new, unseen context values, e.g., an age value not encountered in training.
\section{Problem Setup and Background}
\label{gen_inst} 

Let $X \subseteq \mathbb{R}^D$, $Z \subseteq \mathbb{R}^L$ and $Y \subseteq \mathbb{R}$ be the random variables corresponding to the input explanatory features, input contextual variables, and output prediction value, respectively. Given the realizations of these random variables from some unknown joint data distribution $P_{{X},Z,{Y}}$, the goal of embedded contextual feature selection methods is to achieve the following simultaneously: 1) construct a hyper-model $h_{\bm{\phi}}:\mathbb{R}^L \rightarrow \{0,1\}^D$ that selects a subset of explanatory features $X_{\cal{S}}$ as a function of context $\myvec{z}$; and 2) construct a model $f_{\myvec{\theta}} :\mathbb{R}^D \rightarrow \mathbb{R}$ that predicts the response $Y$ based on these selected features. 
Given a loss function $L$, we solve for the parameters $\myth$ and $\myvec{\phi}$ of the risk minimization problem
\begin{align} \label{eq:optim}
    R(\myvec{\theta}, \myvec{\phi}) = \mathbb{E}_{{X},Z,{Y}} [L(f_{\theta}(\myvec{x} \odot s(\myvec{z})), y)],
\end{align}
where $s(\myvec{z})=h_{\bm{\phi}}(\myvec{z})$ and $\myvec{x},\myvec{z}$ and $\myvec{y}$ represent a realization of the random variables $X,Z$ and $Y$ following a data distribution $P_{X,Z,Y}$, the feature selection vector, given by a vector of indicator variables for the set $\cal{S}$, is the output of the hypernetwork $s(\myvec{z}) = h_{\bm{\phi}}(\myvec{z})= \{0, 1\}^D$, and $\odot$ denotes the point-wise product.

Throughout this paper, bold lowercase letters denote vectors, e.g., $\myvec{x}$, while scalars are represented by unbolded lowercase letters. Elements within a vector are indicated by subscripts, such as $x_1, x_2, \ldots, x_n$. For a function $f$ that maps a vector $\myvec{x}$ to another vector, $f(\myvec{x})$, the $i^{th}$ element of the resultant vector is denoted as $f_i(\myvec{x})$. \myvec{0} and \myvec{1} indicate a vector of all zeros and all ones, respectively. $\myvec{I}$ represents identity matrix. \(d \in [D]\) is shorthand for \(d\) ranging from 1 to \(D\), inclusive.

\section{Conditional STG}
The risk minimization in (\ref{eq:optim}) often includes an additional constraint to induce sparsity on the feature selection, for example, $\|\mys\|_0 \leq D$, which reduces the number of selected features and enhances interpretability. Model interpretability is crucial for understanding complex relations between input features and the predicted target in applications such as health care~\cite{liu2019read}, shopping~\cite{baltrunas2009context}, and psychology~\cite{everaert2022features}.  

In practice, $\ell_0$ regularization is computationally challenging, especially for high-dimensional data, and cannot be integrated into gradient descent-based optimization commonly used in deep networks. Our proposed solution is a probabilistic and computationally efficient method that facilitates contextual feature selection. This is achieved by applying $\ell_0$ regularization to stochastic gates, which mask the input feature conditioned on the context.

First, we introduce conditional Bernoulli gates corresponding to each of the $D$ features as a probabilistic extension of contextual feature selection. 
Expressly, we assume that the probability of an individual explanatory feature being selected, given contextual features $\myvec{z}$, follows a Bernoulli distribution independent of the probability of selecting other explanatory features. 
Let ${S' \vert Z}$ be a conditional random vector that represents these independent Bernoulli gates, with $\prob(S'_{d} = 1 \vert Z=\myvec{z})=\pi_d(\myvec{z})$ for $d \in [D]$, and let $s'(\myvec{z})={s'|\myvec{z}}$ be a realization of the random vector $S'|Z$. We term these conditional-Stochastic Gates (c-STG), as the parameter of the distribution of each gate is conditional on the context variables $\myvec{z}$.

The hypernetwork $h'_{\bm{\phi}}$ learns a parametric function mapping the contextual variable $\myvec{z}$ to the conditional probability, i.e., the parameter of the Bernoulli distribution. 
The task of contextual variable selection then boils down to learning the parameters $\pi(\myvec{z})=h'_{\bm{\phi}}(\myvec{z})$ of this conditional distribution, which spans over a continuous space $[0,1]^D$, instead of a discrete set $\{0,1\}^D$, as in Eq.~\eqref{eq:optim}. 
Let $h'_{\bm{\phi}}: \mathbb{R}^L \to [0,1]^D$ be the function that maps the contextual information to the parameters of the probability distribution. This reformulates the regularized version of the risk in Eq.~\eqref{eq:optim} to 

\begin{equation}
\label{eq:bern_risk}
   \hat{R}({\myvec{\theta}}, \myvec{\phi}) = \hat{\mathbb{E}}_{{X,Z},{Y} } \mathbb{E}_{S'\vert Z}  [  L(f_\myvec{\theta}(\myvec{x} \odot {s'(\myvec{z})}), y) \\  + \lambda  ||{s'(\myvec{z})}||_0   ], 
\end{equation}

where the parameters of the distribution $S'\vert Z$, $\pi(\myvec{z})$, are learnt by the hypernetwork, $h'_{\bm{\phi}}(\myvec{z})$. $\hat{\mathbb{E}}_{X,Z,Y}$ represents the empirical expectation over the observations $X,Z$ and $Y$ and $\mathbb{E}_{S'\vert Z}||{s'(\myvec{z})}||_0=\sum_{d=1}^D \pi_d(\myvec{z})$, the sum of Bernoulli parameters. 
Constraining $\pi_d(\myvec{z})$ to $\{0,1\}$ makes this equivalent to the cardinality-constrained version of Eq.~\eqref{eq:optim}
, with a regularized penalty on cardinality rather than an explicit constraint. Moreover, this probabilistic formulation converts the combinatorial search to a search over the space of Bernoulli distribution parameters. Thus, the problem of feature selection translates to finding $\myvec{\theta}^*$ and $\bm{\phi}^*$ that minimize the empirical risk based on the formulation in Eq.~\eqref{eq:bern_risk}.

\begin{algorithm}[t]
\caption{Weighted c-STG}
\begin{flushleft}
\textbf{Input:}
\begin{tabular}[t]{ll}
$\myvec{x}^{(k)} \in \mathbb{R}^D,\myvec{z}^{(k)} \in \mathbb{R}^L,y^{(k)} \in \mathbb{R}$, for $k=[K]$ \\
\end{tabular} \\
\textbf{Output:} Trained models $f_{\bm{\theta}}$ and $\tilde{h}_{\bm{\phi}}$.
\end{flushleft}
\begin{algorithmic}[1]
\State \textit{Initialize: $\myvec{\theta}$ and $\bm{\phi}$ using Xavier initialization.}
\State \textbf{while} { model not converged} \textbf{do}
\State  \hspace{\algorithmicindent} \textit{Forward Pass:} 
\State  \hspace{\algorithmicindent} for {$k=1$ to $K$ }
\State  \hspace{\algorithmicindent} \hspace{\algorithmicindent} $\mu(\myvec{z}^{(k)}),w(\myvec{z}^{(k)}) = \widetilde{h}_{\bm{\phi}}(\myvec{z}^{(k)}) $ 
\State \hspace{\algorithmicindent} \hspace{\algorithmicindent} $\epsilon \sim \mathcal{N}(0,\sigma^2 \myvec{I})$ 
\State  \hspace{\algorithmicindent} \hspace{\algorithmicindent} $\widetilde{s}(\myvec{z}^{(k)}) = \max(\myvec{0},\min(\myvec{1},\mu(\myvec{z}
^{(k)})+\epsilon))$, \quad where $\min$ and $\max$ are applied elementwise.
\State  \hspace{\algorithmicindent} \hspace{\algorithmicindent} $\hat{y}^{(k)} = f_{\bm{\theta}}(\myvec{x}^{(k)} \odot \widetilde{s}(\myvec{z}^{(k)}) \odot w(\myvec{z}^{(k)}))$ \label{line:yhat}
\State \hspace{\algorithmicindent} $\hat{R}(\bm{\theta},\bm{\phi})\!=\!\!\sum\limits_{k=1}^K \!\left[ L(\hat{y}^{(k)}, y^{(k)})\!+\!\lambda \sum\limits_{d=1}^D \Phi\left(\frac{\mu_d(\bm{z}^{(k)})}{\sigma}\right) \right]$
\State \hspace{\algorithmicindent} \textit{Back Propagation:}
\State  \hspace{\algorithmicindent} \textit{Update $\myvec{\theta}$:} $\myvec{\theta} \leftarrow \myvec{\theta} - \eta \frac{\partial \hat{R}({\bm{\theta}}, {\bm{\phi}})}{\partial {\bm{\theta}}}$ 
\State  \hspace{\algorithmicindent} \textit{Update $\myvec{\phi}$:} $\myvec{\phi} \leftarrow \myvec{\phi} - \eta \frac{\partial \hat{R}({\bm{\theta}}, {\bm{\phi}})}{\partial {\bm{\phi}}}$ 

\end{algorithmic}
\label{alg}
\end{algorithm}

\subsection{Bernoulli Continuous Relaxation for Contextual Feature Selection}
\label{sec:cr}

Incorporating discrete random variables into a differentiable loss function to retain informative data features is appealing, but discrete variable gradient estimates often have high variance \cite{he2004locality}. Consequently, continuous approximations of discrete variables have been proposed \cite{maddison2016concrete,jang2016categorical}. A stable approach for continuous relaxation uses the Gaussian distribution, more consistent in feature selection than Gumbel-softmax techniques like concrete and hard concrete \cite{jang2016categorical}, which can lead to high variance in approximating Bernoulli variables \cite{yamada2020feature,jana2021support}. Such relaxations \cite{louizos2017learning} are applied in various areas, including discrete softmax activations \cite{jang2016categorical}, feature selection \cite{yamada2020feature,lindenbaum2021differentiable,shaham2022deep}, and sparsification \cite{lindenbauml0,yariv}. We utilize a Gaussian-based relaxation for Bernoulli variables, termed Stochastic Gates (STG) \cite{yamada2020feature}, differentiated using the reparameterization trick \cite{reparameterization1,reparameterization2}.

The Gaussian-based continuous relaxation for the Bernoulli variable is defined as $\widetilde{s}_d(\myvec{z}) = \max(0,\min(1, \mu_d(\myvec{z}) + \epsilon_d)$, where $\epsilon_d$ is drawn from a normal distribution $\mathcal{N}(0,\sigma^2)$, with $\sigma$ fixed throughout training. 
Unlike the ReLU function, which only clips the negative values to zero, the mean-shifted Gaussian variable clips both the positive and negative values; therefore, it accounts for the binary nature of the original random variable.
Here, we learn $\mu_d$ as a parametric function of the contextual variables $\myvec{z}$.
Thus, the hypernetwork $\widetilde{h}_{\bm{\phi}}$ aims to learn the parameters of the relaxed-continuous distribution as a function of the context variables $\myvec{z}$ instead of learning the original discrete distribution. Our objective as a minimization of the empirical risk $\hat{R}({\bm{\theta}}, {\bm{\phi}})$ is as follows:
\begin{align}
    \min_{{\bm{\theta}, \bm{\phi}}} \hat{\mathbb{E}}_{{X,Z},{Y}} \mathbb{E}_{{\widetilde{S} \vert Z}} \big[ 
    L(f_{\bm{\theta}}({\myvec{x}}\odot {\widetilde{s}(\myvec{z})}), y) 
    + \lambda  ||{\widetilde{s}(\myvec{z})}||_0  \big], & \label{eq:risk_p} 
\end{align}
where $\widetilde{S}|Z$ is a random vector with $D$ independent variables $\widetilde{\myvec{s}}_d(\myvec{z})=\widetilde{\myvec{s}}\vert \myvec{z}$ for $d \in [D]$ and the parameters of the distribution $\widetilde{S}\vert Z$, $\pi(\myvec{z})$, are learnt by the hypernetwork, $\widetilde{h}_{\bm{\phi}}({\myvec{z}})$. The regularization term can be further simplified to

\begin{align}
    \mathbb{E}_{\widetilde{S} \vert Z}  ||\widetilde{s}(\myvec{z})||_0  
    = \sum_{i=1}^D \prob(\widetilde{s}_d(\myvec{z}) > 0)  = \sum_{d=1}^D \Phi ( \frac{\mu_d(\myvec{z})}{\sigma} ),  
    \label{eq:reg_term}
\end{align}

 where $\Phi$ is the standard Gaussian CDF. The term \eqref{eq:reg_term} penalizes open gates so that gates corresponding to features that are not useful for prediction are encouraged to transition into a closed state (which is the case for small $\mu_d(\myvec{z})$). Hence, we perform a context-specific feature selection strategy by inducing sparsity through the empirical mean of the regularization term (\ref{eq:reg_term}) over multiple realizations of $Z$. This enables us to select distinct informative features for different contexts while maintaining the sparsity.

 In practice, we consider the function class for $\widetilde{h}_{\bm{\phi}}$ and $f_{\bm{\theta}}$ to be a class of neural networks parameterized by ${\bm{\phi}}$ and ${\bm{\theta}}$, respectively. To optimize for these parameters, we use a Monte Carlo sampling gradient estimator of (\ref{eq:risk_p}), which gives

\begin{equation}
\begin{split}
\frac{\partial \hat{R}({\bm{\theta}}, {\bm{\phi}})}{\partial {\bm{\theta}}}  = \frac{1}{K} \sum_{k=1}^K &\frac{\partial }{\partial {\bm{\theta}}} L(f_{\bm{\theta}}({{\myvec{x}^{(k)}}}\odot {\widetilde{s}(\myvec{z}^{(k)})},y^{(k)}) 
\\
    \frac{\partial \hat{R}({\bm{\theta}}, {\bm{\phi}})}{\partial {\bm{\phi}}} = \frac{1}{K} \sum_{k=1}^{K} &\frac{\partial}{\partial \phi} \mathcal{L}(f_{\theta} (x^{(k)} \odot \widetilde{s}(z^{(k)})), y^{(k)}) + \lambda \frac{\partial}{\partial \phi} \Phi\left(\frac{\widetilde{h}_{\phi}(z^{(k)})}{\sigma}\right) 
\end{split}
\label{eq:diff_phi1}
\end{equation}

where $K$ is the number of Monte Carlo samples (corresponds to the batch size). 
Our methodology is summarized in Alg.
~\ref{alg} and illustrated in Fig.~\ref{fig:blockdiagram}.
Note for c-STG, the explanatory features are masked by the feature gates, $\widetilde{s}(\myvec{z})$, and fed into the prediction model $\hat{y}^{(k)} = f_{\bm{\theta}}(\myvec{x}^{(k)} \odot \widetilde{s}(\myvec{z}^{(k)}) )$. 

In the initial training phase $\bm{\phi}$, all gates should have an equal probability of being open or closed. We set $\mu_d(\myvec{z})=0.5$ $\forall$ $d \in [D]$ so that all gates approximate a ``fair" Bernoulli variable. The initialization of the hyper-model's $\bm{\phi}$ using Xavier initialization~\cite{glorot2010understanding} and a Sigmoid activation function in the final layer of $\widetilde{h}_{\bm{\phi}}$ ensures that the means ($\mu_d(\myvec{z})$) are centered around 0.5, $\forall$ $\myvec{z}$, early in the training phase. It is worth noting that we need the noise term only during the training phase.

\subsection{Theoretical Analysis}
We conduct a thorough theoretical analysis to establish the equivalence between our probabilistic formulation, as represented by Eq.\ref{eq:bern_risk}, and the original NP-hard contextual variable selection problem defined in Eq.\ref{eq:optim}. Additionally, we provide proof demonstrating that c-STG achieves a lower risk than STG. Furthermore, we extend our analysis to a linear regression scenario, where we demonstrate the main advantage of c-STG over STG. While STG selects features with consistent significance across contextual variables on average, c-STG adapts the feature selection process according to the contextual variables, thus effectively learning the optimal feature selection as a function of these variables. 

\begin{restatable}{thm}{probrelaxation}
Let $s^*(\myvec{z})$ and ${s'}^{*}(\myvec{z})$ represent the optimal feature selection functions in Eq.~\eqref{eq:optim} and its corresponding probabilistic formulation in Eq.~\eqref{eq:bern_risk} respectively. Then $s^*(\myvec{z})=s'^{*}(\myvec{z})$. 
\label{thm:prob_relaxation}
\end{restatable}

This theorem suggests that a deterministic search for feature selection can be transitioned to a probabilistic approach. We use the universal function approximators, deep networks, to learn the function $s'^{*}(\myvec{z})$.
In subsequent theorems, we contrast c-STG's and STG's performance and feature selection abilities.
In \cite{yamada2020feature}, the optimization problem of the global feature selection using STG is given by
\begin{equation}
\label{eq:bern_risk_STG}
   \min \hat{\mathbb{E}}_{{X},{Y} } \mathbb{E}_{S'} \left [ L(f_{\theta}({\myvec{x}} \odot {\myvec{s'}}), y) + \lambda  ||{\myvec{s'}}||_0  \right ] ,
\end{equation} 
where $S'$ is a random vector that represents independent Bernoulli gates, $\prob({S'_{d}} = 1)=\pi_d$ for $d \in [D]$, and ${s'_d}$ denotes the realization of the random vector $S'_d$. 
Note that $S'$ is similar to $S'(\myvec{z})$ except that the latter is a function of $\myvec{z}$, and the former is constant for all $\myvec{z}$.
Thus, their approach optimizes for a fixed feature selection independent of contextual information by promoting sparsity in feature selection through the regularization term $\mathbb{E}_{{{\widetilde{S}}}} \left [ ||{{\widetilde{\myvec{s}}}}||_0  \right ]$.
In contrast, we perform context-specific feature selection by maximizing the sparsity through the empirical mean of Eq.~\eqref{eq:reg_term} across various realizations of $Z$.
The following theorem draws a connection between their empirical risk minimizations.
\begin{restatable}{thm}{lowriskcSTG}
    c-STG attains an optimal risk lower or equal to the risk attained by STG.  
    \label{thm:low_risk_cSTG}
\end{restatable}
 
Through the following Theorems~\ref{thm:stg_relation_pi} and \ref{thm:stg_relation_mu}, we further emphasize the advantage of c-STG over STG. 
\begin{restatable}{thm}{stgrelationpi}\label{thm:stg_relation_pi}
In a linear regression setup, the relationship between the optimal parameters of the conditional Bernoulli stochastic gates, $\pi^*(\myvec{z})$, and the optimal parameters of the non-conditional Bernoulli stochastic gates $\myvec{\pi}^*_\textrm{stg}$ is given by
\begin{equation}
    \myvec{\pi}^*_\textrm{stg} = E_{Z} [\pi^*(\myvec{z})].
\end{equation}
\end{restatable}
\begin{restatable}{thm}{stgrelationmu}\label{thm:stg_relation_mu}
In a linear regression setup, the relationship between the optimal parameters of the conditional Gaussian stochastic gates, $\mu^*(\myvec{z})$, and the optimal parameters of the non-conditional Gaussian stochastic gates $\myvec{\mu}^*_\textrm{stg}$ is given by
\begin{equation}
    \myvec{\mu}^*_\textrm{stg} = E_{Z} [\mu^*(\myvec{z})].
\end{equation}
\end{restatable}
The theorems illustrate that c-STG offers an enhanced feature selection resolution by pinpointing features crucial to specific contexts. This stands in contrast to STG, which tends to select features based on their average importance across various contexts. Theorem \ref{thm:stg_relation_pi} delineates this difference in the realm of discrete probability spaces, whereas Theorem \ref{thm:stg_relation_mu} addresses the continuous probability spaces.
The proofs of all theorems are provided in the appendix.

\subsection{Weighted Conditional STG}
\label{sec:wc-stg}
We extend c-STG to weighted conditional-STG (weighted c-STG) to determine and quantify the significance of the features identified by the conditional stochastic gates. 
To achieve this, we integrate an additional layer in our model that maps the hypernetwork's penultimate layer, $h_L(\myvec{z})$, to a weight vector $w(\myvec{z}) = Wh_L(\myvec{z}) + b$, as depicted in Figure~\ref{fig:blockdiagram} (green circles). 
Explanatory features $\myvec{x}$ are masked by the feature selection output $\widetilde{s}(\myvec{z})$ and weighted by $w(\myvec{z})$, then fed into the prediction model for task execution. 
The hypernetwork and prediction network parameters are learned using back propagation, as detailed in Algorithm~\ref{alg}.
\begin{restatable}{thm}{lowriskwcSTG}
    \label{thm:low_risk_wcSTG}
    Weighted c-STG attains an optimal risk lower or equal to the risk attained by c-STG.  
\end{restatable}
\begin{figure}[t]
    \centering
    \vskip -0.05in
   \includegraphics[width=0.8\linewidth]{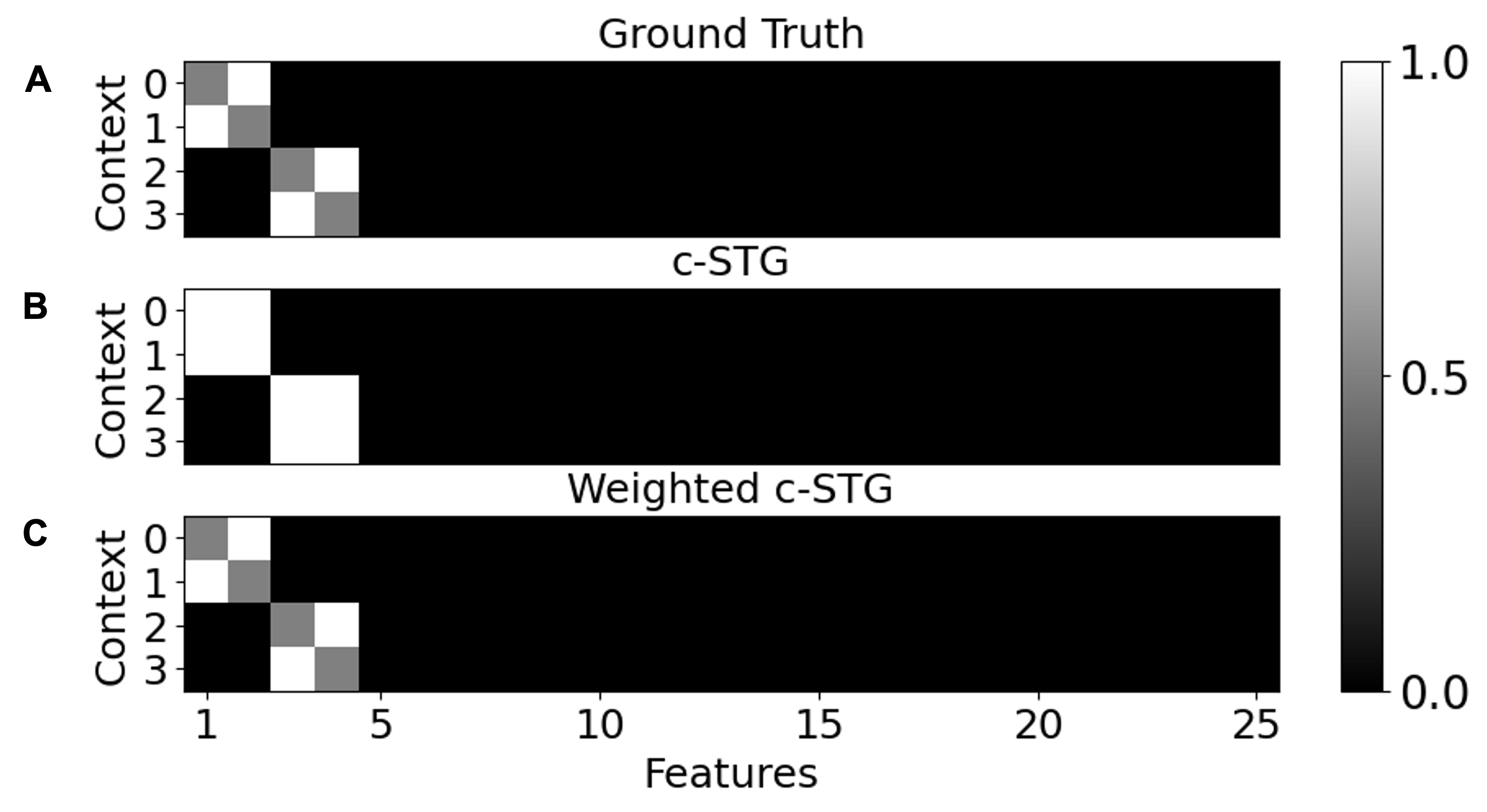}
   \vskip -0.1in
    \caption{XOR2. (A) Ground truth feature significance as a function of context, $z$. Feature gates for c-STG (B) and weighted c-STG (C).}
    \vskip -0.1 in
    \label{fig:nonl_combo}
\end{figure}

\section{Experiments\footnotemark\label{sec:experiments}}

\footnotetext{Code for c-STG is available in \url{https://github.com/Mishne-Lab/Conditional-STG}.}

We evaluate c-STG and weighted c-STG against multiple benchmarks: 1) context-specific techniques Contextual LASSO and CEN; 2) the sample-specific method INVASE~\cite{yoon2018invase}; 3) an active feature selection approach: AFS~\cite{covert2023learning}; 4) population-level methods like LASSO and STG; and 5) a prediction model without any feature selection.
We train the population-level methods and prediction model without any feature selection using either 1) only the explanatory features as input or 2) concatenating these features with the context variables (referred to as `with context'). 
See supplementary material (Appendix \ref{sec:hyperparameters}) for implementation details on the hyper-parameters of all methods. 
We report the performance of all methods in Table~\ref{tab:results} and their corresponding number of selected features in (Table~\ref{tab:num_features}) for four datasets, demonstrating c-STG and weighted c-STG outperform competing methods.

\noindent \textbf{Simulated Datasets: }
First, we use synthetic data to validate that our method can identify context-specific informative features while learning non-linear prediction functions. 
Following synthetic examples in \cite{yamada2020feature,yang2022locally}, we design a nonlinear moving XOR dataset (XOR1), described in the appendix, and a nonlinear weighted moving XOR (XOR2) as follows.
We generate a data matrix ${X}$ with $1000$ samples and $25$ features, where each entry is sampled from a normal distribution. The contextual variable $\myvec{z}$ is sampled uniformly from $\{0,1,2,3\}$. The prediction variable $\textnormal{y}$ is generated as
\begin{align*}
    \textbf{XOR2:} \hspace{0.2cm} 
    \textnormal{y}(\myvec{x};\myvec{z}) &=
    \begin{cases}
        ReLU(0.5{x}_1 + {x}_2),& \text{if } z = 0,\\ 
       ReLU({x}_1 + 0.5{x}_2) ,& \text{if } z = 1,\\
        ReLU(0.5{x}_3 + {x}_4) ,& \text{if } z = 2,\\
        ReLU({x}_3 + 0.5{x}_4) ,& \text{if } z = 3.\\
    \end{cases} . 
\end{align*}
Based on the value of $\myvec{z}$, the response variable $\textnormal{y}$ for different samples will have different feature significance. Fig.~\ref{fig:nonl_combo} shows that c-STG correctly recovers which features are significant for prediction while weighted c-STG recovers context-dependent importance. 
Furthermore, we introduce a third synthetic example (XOR3) in the appendix, which demonstrates c-STG's advantage over contextual LASSO, particularly by showcasing how the latter is prone to shrinkage problems in a linear context.

\begin{table*}[t]
\centering
 \caption{Comparison of feature selection. \textbf{Bold} indicates best performance and \underline{underline} indicates second-best. }
\begin{tabular}{c|c||c||c||c||c|}
\cline{2-6} & \multicolumn{1}{c||}{\textbf{XOR1}}
 & \multicolumn{1}{c||}{\textbf{XOR2}} & \multicolumn{1}{c||}{\textbf{MNIST}}& \multicolumn{1}{c||}{\textbf{Heart disease}}  & \multicolumn{1}{c|}{\textbf{Housing}}                        \\ \cline{2-6}
 & Accuracy ($\%$ ) & $R^2$ score   & Accuracy ($\%$ )  & Accuracy ($\%$) & $R^2$ score   \\ \hline
\multicolumn{1}{|c|}{LASSO}                 & 50.09 (1.23) & 0.3109(0.0175)
 & 83.87 (0.05)   &  83.50 (3.69)                       & 0.2161 (0.0014)                             \\ 
\multicolumn{1}{|c|}{with context}   & 50.05 (1.09) & 0.3105(0.0170) & 83.88 (0.04)   &  83.67 (3.48)                       & 0.2272 (0.0016)                              \\ \hline
\multicolumn{1}{|c|}{Prediction model}            & 70.74 (17.72)  & 0.3409 (0.0192) & 98.11 (0.05)    &86.53 (1.51)                       & 0.2022 (0.0361)                      \\ 
\multicolumn{1}{|c|}{with context} & {{ 71.93}} (11.88) & 0.3382 (0.0167) & 98.21 (0.07)      & 83.92 (5.52)                    & 0.2132 (0.0381)                      
                   \\ \hline
                   
\multicolumn{1}{|c|}{STG }                  & 73.98 (0.69)  & 0.3516 (0.0287) & {98.55} (0.08)   & 86.53 (2.11)                & 0.2139 (0.0023)                        \\ 
\multicolumn{1}{|c|}{with context}     & 74.22 (1.25) & 0.3487 (0.0269) & 98.34 (0.07)    & \ul{87.88} (1.95)                      & { 0.2234} (0.0022)                      \\ \hline
\multicolumn{1}{|c|}{CEN}                  & 50.08 (0.67) & 0.5186 (0.0126)
 & 87.66 (0.24)   & 84.17 (5.69)
                  & 0.2561 (0.0002)                      \\ \hline

\multicolumn{1}{|c|}{Contextual LASSO }                  & 50.60 (0.61) & 0.7316 (0.0090)

 & 97.17 (0.04) & 83.19 (5.32) 
                 & \ul{0.4616} (0.0033)
                       
    \\ \hline
\multicolumn{1}{|c|}{AFS }                  & \ul{78.95} (19.91) & 0.7572 (0.0187)
 & 94.83 (0.84)  & 86.44 (1.07) 
                 & 0.3963 (0.0092)                  
    \\ \hline

\multicolumn{1}{|c|}{INVASE }                  & 52.36(1.70) & 0.4627(0.0194)
 & 88.06(2.30) & 87.22(2.20) 
 & 0.3335 (0.0203)
 \\ \hline

\multicolumn{1}{|c|}{c-STG (Ours) }           & \textbf{100} (0)  &  \ul{0.8739} (0.0107)    & \ul{98.66} (0.05)        &  {87.55} (3.41)          & { {0.3976}} (0.0082)           \\ 
\hline

\multicolumn{1}{|c|}{Weighted c-STG (Ours) }           & \textbf{100} (0)    & \textbf{0.9956} (0.0008)
  & \textbf{98.69} (0.06)        & \textbf {89.23} (1.72)          & {\textbf {0.5308}} (0.0052)           \\ 
\hline

\end{tabular}
\label{tab:results}
\end{table*}

\begin{table*}[t]
\centering
 \caption{Comparison of a number of features selected on XOR, MNIST, Housing, and heart disease datasets. \textbf{Bold} indicates least number of selected features and \underline{underline} indicates second-best. }
\small
\begin{tabular}{c|c||c||c||c||c|}
\cline{2-6}
 & \multicolumn{1}{c||}{\textbf{XOR1}} & \multicolumn{1}{c||}{\textbf{XOR2}} 
                                            & \multicolumn{1}{c||}{\textbf{MNIST}} & \multicolumn{1}{c|}{\textbf{Heart disease}}& \multicolumn{1} {c|}{\textbf{Housing}}                         \\ \cline{1-6}
                                            
\multicolumn{1}{|c|}{LASSO}                 & 18.80 (0.87)& 14.20 (2.04)
& 621.70 (4.77)  & 13.6 (1.11)
 & 8.00 (0.00)
       \\ 

\multicolumn{1}{|c|}{with context}                 & 21.50 (0.92)& 14.60 (1.85)
& 627.5 (3.26) & 16.90 (0.70)
 & 10.00 (0.00) 

       \\ \hline

\multicolumn{1}{|c|}{STG}                 & 6.60 (1.50) & 5.60 (1.02)
& 585.40 (3.77) & 16.60 (2.65) & 8.00 (0.00) 

       \\ 

\multicolumn{1}{|c|}{with context}                 & {\ul{4.20}} (0.40)
 & 4.20 (0.40) &\textbf{148.80} (2.64) & 18.00 (4.15) & 8.00 (0.00)

       \\ \hline

\multicolumn{1}{|c|}{Contextual LASSO}                 & 4.60 ± 1.93
 & 2.75 (0.93)
& 357.72 (8.88) & 11.79 (3.66) & 5.43 ( 0.29) 

       \\ \hline

\multicolumn{1}{|c|}{AFS}                 & 5.40 (4.27)
 & \textbf{2.00} (0.00) & 290.00 (20.00) & \ul{8.00} (5.48)
 & \textbf{2.70} (0.64) 

       \\ \hline

\multicolumn{1}{|c|}{INVASE}                 & 9.85(1.47)
 & 13.04 (0.09) & 376.49 (12.30) & 9.12 (1.85) & 4.80 (0.16)

       \\ \hline

\multicolumn{1}{|c|}{c-STG (Ours)}                 & \textbf{3.00} (0.00)
 & \textbf{2.00} (0.00)
& 247.25 (10.84)& 9.69 (0.81) & \textbf{2.70} (0.42) 
\\ \hline

\multicolumn{1}{|c|}{Weighted c-STG (Ours)}                 & \textbf{3.00} (0.00)
 & \textbf{2.00} (0.00)
&\ul {217.02} (2.87)& \textbf{6.50} (0.50) & \ul{4.36} (0.30) 
\\ \hline

\end{tabular}
       \label{tab:num_features}
\end{table*}

\noindent \textbf{MNIST: } We apply our proposed method for image classification using a subset of the MNIST dataset, including digit images of 4 and 9. To create a contextually diverse set of input images, we rotate each original image by intervals of 45 degrees, resulting in eight distinct images per original image. 
In this example, the goal is to identify which pixels most effectively differentiate between the digits per rotation angle.
Applying c-STG or weighted c-STG where rotation serves as context results in a test accuracy of $~98\%$.
In Fig.\ref{fig:mnist}, we contrast STG with c-STG, illustrating the superiority of context-specific feature selection over population-level analysis. By comparing separate STG models for distinct categorical contexts against a unified c-STG model, we find that c-STG leads to higher accuracy with fewer samples~(Fig.\ref{fig:num_samples_accuracy} in the appendix).

\begin{figure}[t]

    \centering
    \includegraphics[width=0.6\linewidth]{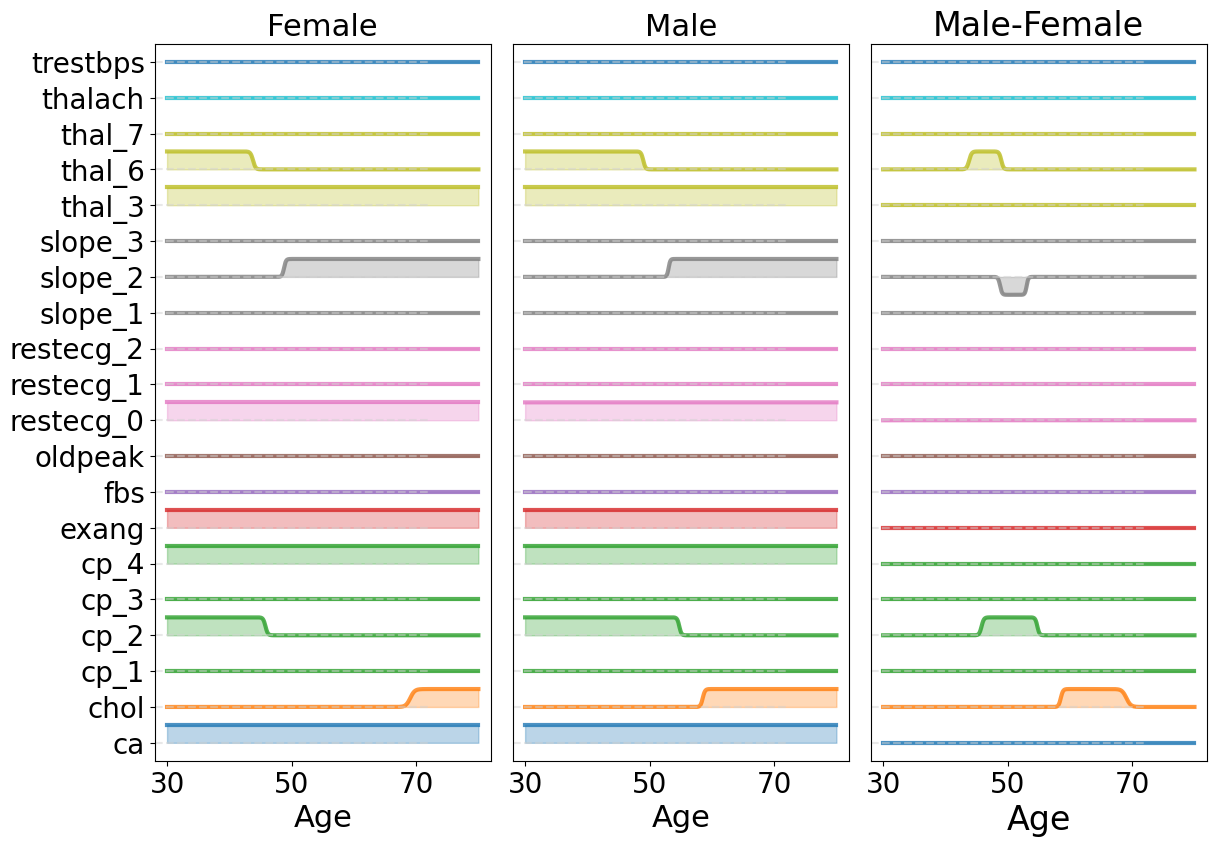}
    \caption{Heart disease. Feature selection gates $\mu(z)$ for each input feature as a function of context\textemdash age and gender (left - females and center - males)\textemdash produced by weighted c-STG. The difference in the c-STG values between males and females indicates gender-specific informative features as a function of age.}
    \label{fig:heartdisease}
\end{figure}

\noindent \textbf{Heart disease dataset: }
    
We now focus on medical data, specifically, the heart disease dataset from UCI ML repository~\cite{misc_heart_disease_45}.
Given features including chest pain type, resting blood pressure, serum cholesterol levels, and more, 
our goal is to understand how age and gender influence these biometrics in relation to the risk of heart disease. 
Age and gender are set as a two-dimensional contextual input to our hypernetwork in this binary classification problem. 
Table~\ref{tab:results} shows the 5-fold cross-validation accuracy where we surpass other methods.

Weighted c-STG analysis (Fig.~\ref{fig:heartdisease}) reveals age and gender-specific feature significance for heart disease. 
For example, cholesterol (`chol') monitoring becomes crucial with increased age as a risk factor for cardiovascular disease. In males, cholesterol's relevance arises around age 50, likely due to the absence of hormone-induced cholesterol regulation. In females, a significant cholesterol risk arises later, past age 70, when they are post-menopause, and estrogen's protective effect has faded.
Similarly, a flat ST segment slope ('slope$\_$2') indicates heart disease risk in females starting at age 45, intensifying post-menopause. For males, `slope$\_$2' becomes significant after 55, showcasing distinct cardiovascular risk patterns between genders.

\noindent \textbf{Housing dataset: }
\begin{figure}
    \centering
    \includegraphics[width=0.6\linewidth]{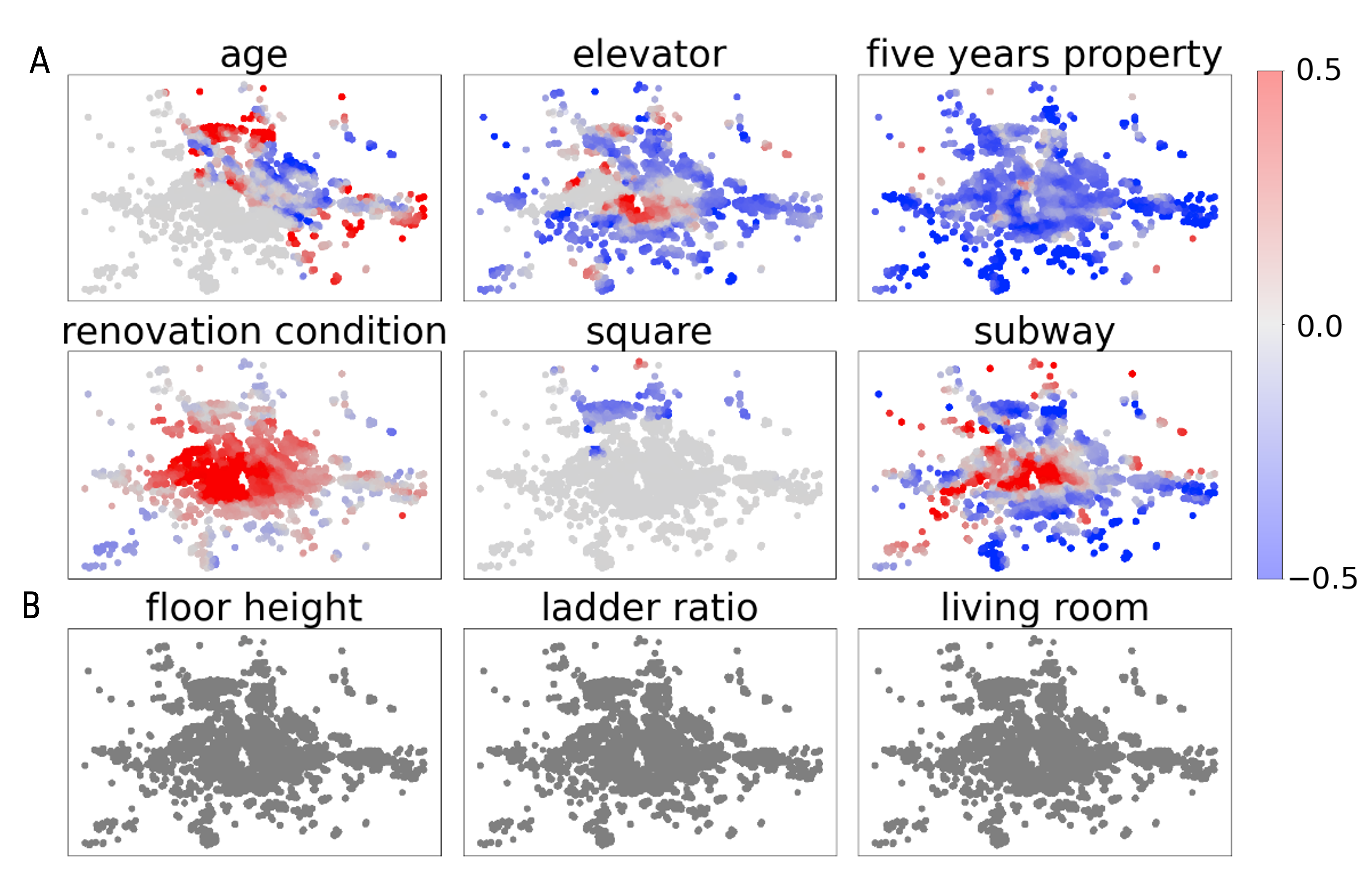} 
    \caption{Housing. Geographic significance of nine housing features according to weighted c-STG analysis ($w(z)\times \widetilde{s}(z)$). The red to blue gradient denotes positive to negative impact, respectively, with gray as neutral. Weighted significant features (A) and unselected features (B).}
    \label{fig:housing}
\end{figure}

In the domain of housing price prediction, understanding how various features, such as the renovation condition or floor height, influence its cost based on the house's location in the city requires context-specific feature selection.
The Housing dataset~\cite{lianjia} includes a set of features such as age, elevator, 
floor height, renovation condition, and the latitude and longitude of the house. We used our proposed approach to predict the price of the house using all the features while considering the location (latitude and longitude) as two-dimensional contextual data. For this dataset, our weighted c-STG outperforms all other models.
Visualizing weighted c-STG feature significance in Fig.~\ref{fig:housing} offers key insights into the Housing dataset. First, feature importance shows a localized pattern, with neighboring locations displaying similar feature significance. Secondly, subway presence in the city center positively impacts house pricing, possibly due to enhanced accessibility and convenience. In contrast, subway presence negatively affects prices in the outskirts, potentially due to increased noise and disruption. Additionally, renovated properties in the city center positively influence pricing, likely because renovations in high-demand urban areas add significant value to properties.
These findings provide valuable insights for real estate companies and policymakers. By understanding which features are more important in which locations, they can make informed decisions regarding housing prices, regulations, and development.

\noindent \textbf{Neuroscience: }
In studying the way the brain encodes behavior, perception, and memory, machine learning models are trained to predict measures of behavior from neuronal activity recordings. Feature selection can provide valuable insights by revealing individual (or subsets of) neurons encoding a particular behavior as a function of task-relevant timing, sensory input, or arousal state. 
In~\cite{levy2020cell}, the authors study a motor task where mice are trained to perform a hand-reach task of a food pellet, given an auditory cue (tone).
They recorded the activity of neurons in layers 2-3 of the primary motor cortex, where each trial was labeled as successful if the animal managed to grab and consume the food pellet. 
Cellular networks in the motor cortex are expected to communicate error signals while acquiring a new motor task to achieve improvement across attempts. 
To test this hypothesis, \cite{levy2020cell} trained a separate SVM binary model to classify trials as success or failure per neuron based on the activity during a short time window. This resulted in training 7866 models (342 neurons $\times$ 23 sliding time windows). The analysis showed that  1) outcome can be consistently decoded from neuronal activity starting 2 seconds after the tone till the end of the trial; 2) 12\% of neurons exhibited prolonged activity starting 2 seconds after the tone, on either success or failure trials, thus ``reporting" trial outcome.
\begin{figure}[t]
    \centering
    \includegraphics[width=0.8\linewidth]{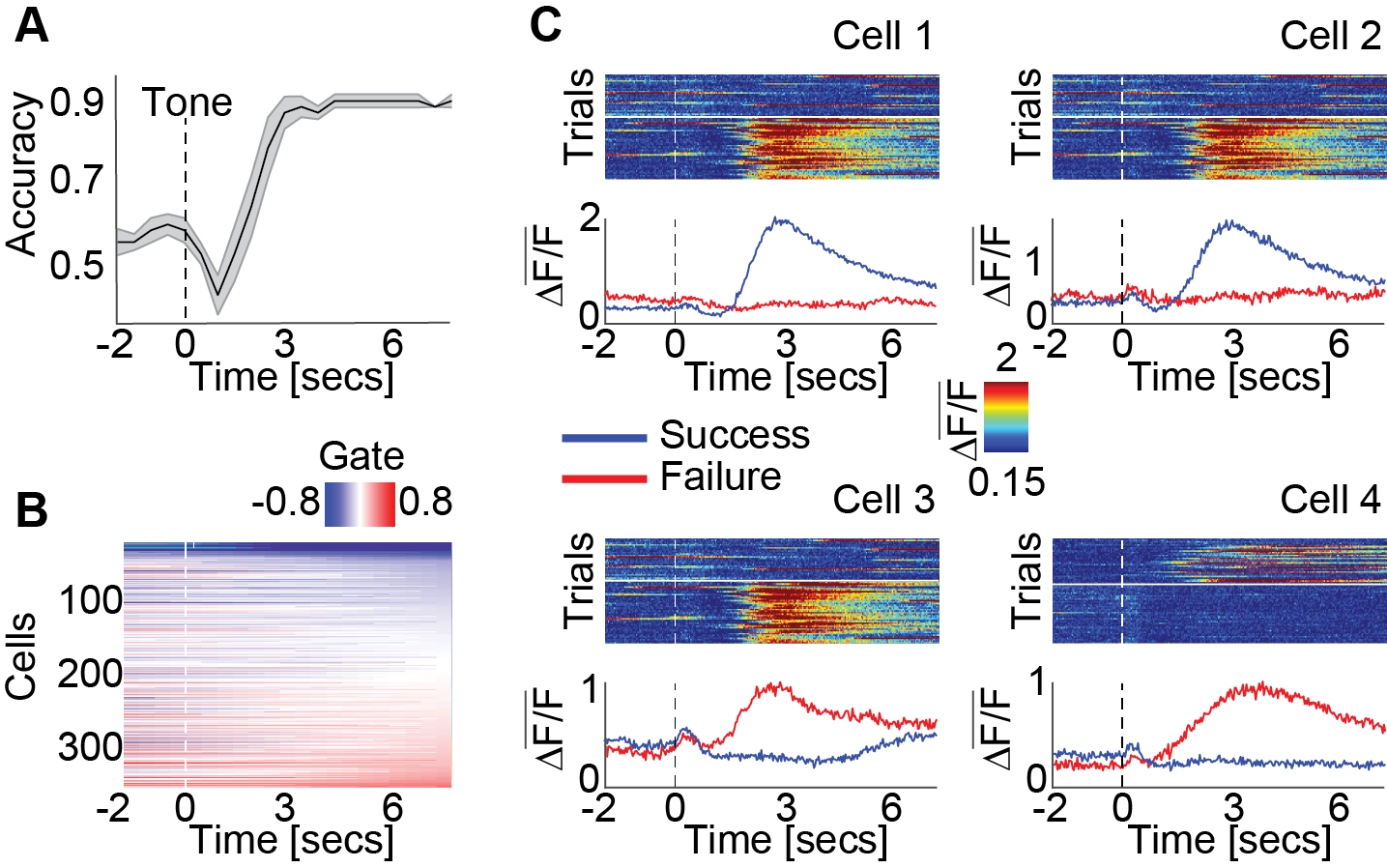}

    \caption{Neuroscience: Context is time. (A) Prediction accuracy vs. time. (B) Weighted gate values for all cells vs. time. (C) Examples of cells selected were for each cell: top - activity vs. time for all successful and failed trials (separated by white horizontal line), bottom - traces of average activity for successful (red) and failed trials (blue).}

    \label{fig:motor23}
\end{figure}

Using a single c-STG model, we reveal the same biological findings.
We set the prediction model to be a binary classifier (success or failure); the explanatory features are the neuronal activity of all neurons in a given time window, and the contextual variable is \textit{time}.
Fig.~\ref{fig:motor23} (A) presents the test accuracy of weighted c-STG as a function of time, indicating that the trial outcome was successfully classified starting from 2 seconds after the tone. The learned weights are presented in (B), where 12\% of cells have absolute weight values that are larger than 0.25. Four example neurons with high absolute weights (C) exhibit prolonged activity only during either success or failure. Overall, weighted c-STG captured the complex dynamics of how outcome is encoded within the ensemble neuronal activity; it was able to detect individual neurons reporting outcome as a function of time using a single model, a biological finding which ~\cite{levy2020cell} used thousands of SVM models to reveal. 

\begin{figure}[t]
    \centering
    \includegraphics[width=0.8\linewidth]{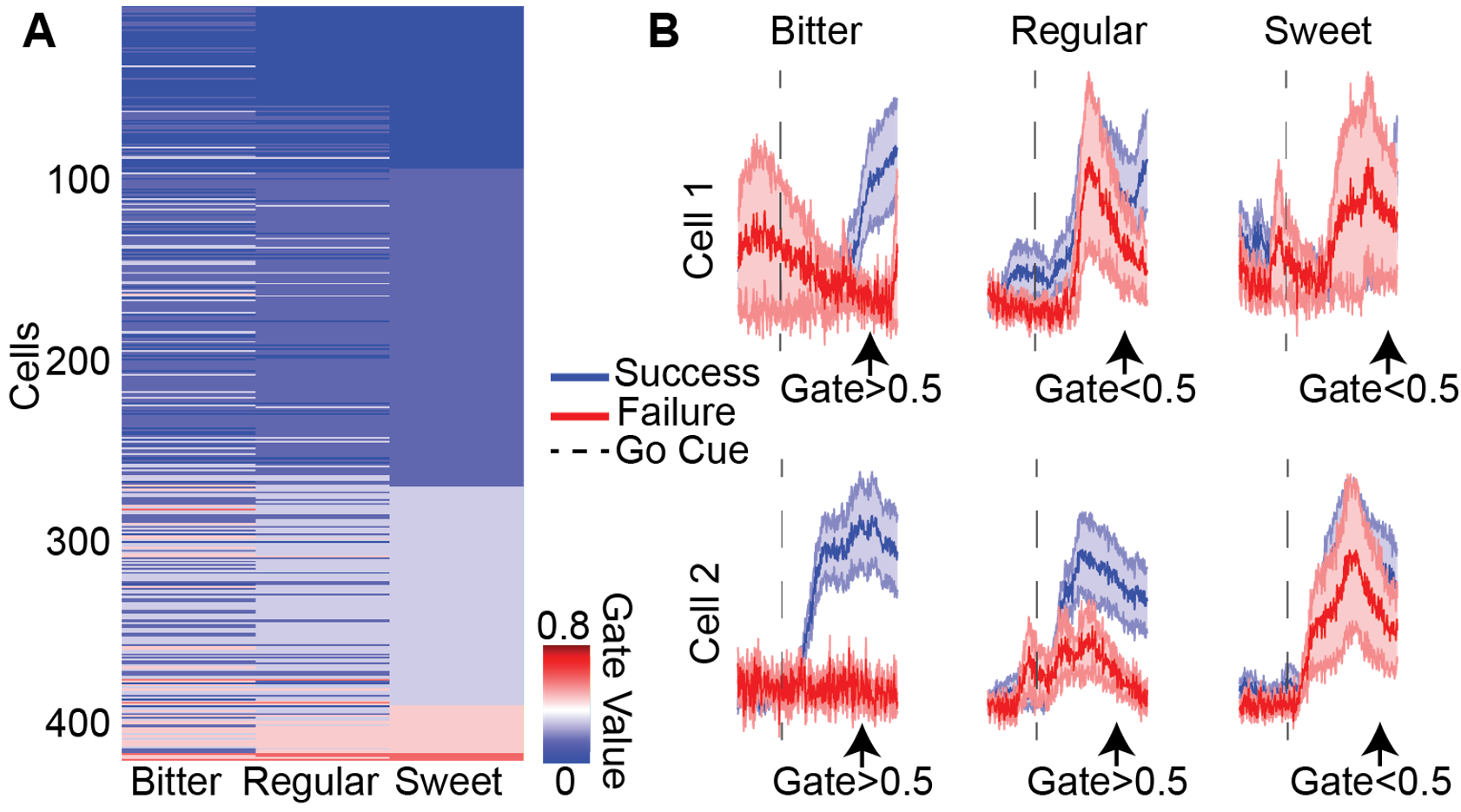}

    \caption{Neuroscience: Context is flavor. (A) c-STG gate values for all cells per flavor. (B) Examples of cells selected by c-STG (gate $>$ 0.5); Averaged activity across trials (mean$\pm$ Standard Error of the Mean) for successful (blue) and failed (red) trials demonstrating differences in outcome encoding across flavors.} 

    \label{fig:flavors}
\end{figure}

Finally, we apply c-STG to another hand-reach study where a mouse was given flavored food pellets: bitter, sweet, or regular (unflavored). We hypothesize that outcome encoding depends on flavor. We train c-STG to classify the outcome based on the activity in the last 4 seconds of each trial, where the contextual variable is \textit{flavor}. Our results in Fig. \ref{fig:flavors} reveal that the cellular network encodes the outcome differently across flavors. We quantify how similar the encoding is by correlating the gate values across flavors and find they are most different for sweet and bitter  (c=0.4), and they are more similar for regular and sweet (c=0.7) than for regular and bitter (c=0.6). Overall, c-STG is able to capture the complexity of outcome encoding by the neuronal network across different contexts.

\section{Conclusion}
We presented an embedded method for context-specific feature selection. We developed conditional stochastic gates based on a hypernetwork to map between contextual variables and the parameters of the conditional distribution
Our c-STG leads to improved accuracy and interpretability by efficiently determining which input features are relevant for prediction using a single trained model for categorical, continuous, and/or multi-dimensional context.
We demonstrate that our method outperforms embedded feature selection methods and reveals insights across multiple domains. 
A limitation of our work is the challenge of parameter tuning, specifically the regularization and cross-validation, which is often necessary to find optimal value. Notably, LASSO has a well-established theory for tuning its $\lambda$ parameter. Developing automated procedures for parameter tuning for c-STG would be an exciting avenue for future research.

\section*{Acknowledgements}
This research is partially supported by a Simons Foundation Pilot Extension Award - 00003245 (G.M. and R.S.) and Israel Science Foundation 1432/19 (J.S.).

\section*{Impact Statement}
The potential societal impact of our work is broad and multifaceted. On a positive note, c-STG's ability to incorporate contextual information into feature selection could impact fields such as healthcare by enabling more accurate patient diagnostics and tailored treatments or urban planning through improved housing and infrastructure development models. However, as with any technological advancement, the potential for misuse exists. For instance, using contextual data like sentiment or mood in social networks could lead to manipulative practices, such as exploiting users' emotional states for targeted content delivery that may not be in their best interest. Similarly, in targeted advertising, c-STG could be employed to differentiate offerings based on sensitive context variables (e.g., location, race), leading to inequitable practices like dynamic pricing or content filtering that exacerbate societal divides.

As researchers, we acknowledge these potential ethical dilemmas and stress the importance of developing and implementing safeguards against misuse. This includes promoting transparency in how models are trained and used, ensuring diverse and inclusive data to prevent bias, and advocating for regulations that protect individuals' rights and privacy. Moreover, it is vital to foster interdisciplinary dialogue between technologists, ethicists, policymakers, and stakeholders to navigate the ethical complexities and societal implications of advanced machine learning technologies.

In conclusion, while c-STG represents a significant leap forward in context-aware machine learning, it is incumbent upon the research community and society at large to carefully consider and address the ethical and societal ramifications of such technologies. By doing so, we can harness the full potential of c-STG and similar innovations for the greater good, ensuring they serve to unite rather than divide, empower rather than exploit, and foster equity rather than exclusion.

\bibliography{main}

\newpage

\appendix
\section{Theoretical Proofs}
\probrelaxation*

\textit{Proof: }
 Firstly, the optimal solution for Eq.~\eqref{eq:optim} lies in the feasible solution space for Eq.~\eqref{eq:bern_risk}. Secondly, Eq.~\eqref{eq:bern_risk} attains its optimal value at the extreme values of the Bernoulli parameters $\pi(\myvec{z})$. We refer the reader to \cite{yin2022probabilistic} for detailed proof.

\lowriskcSTG*
\textit{Proof: } The feasible solution space of STG is contained in the feasible solution space of c-STG as the parameters of the stochastic gates in c-STG can be any function of contextual variables $\myvec{z}$, which includes a constant function, which is the case in STG. Therefore, the optimal risk attained by c-STG must be less than or equal to the optimal risk value attained by STG.

\stgrelationpi*
\textit{Proof: } The proof proceeds in five steps.

 Step 1: When we consider $L(\cdot,\cdot)$ to be a mean squared loss, the solution to the optimization problem in Eq.~\eqref{eq:bern_risk_STG} and Eq.~\eqref{eq:bern_risk} will be given by their respective MMSE (minimum mean squared error) estimates. These are given by,
\begin{equation}\label{eq:mmse_stg}
    f_{\bm{\theta}^{*}}(\myvec{x}\odot \myvec{{s'}}) = E[Y|X=\myvec{x},S'=\myvec{s'}]
\end{equation}
\begin{equation}\label{eq:mmse_cstg}
    f_{\bm{\theta}^{*}}(\myvec{x}\odot {s'(\myvec{z})}) = E[Y|X=\myvec{x},Z=\myvec{z},S'(\myvec{z})=s'(\myvec{z})]
\end{equation}
where the feature selection vector $\myvec{s'}$ and contextual feature selection vector $s'(\myvec{z})$ will be sampled from a constrained space.

Step 2: Relaxing $f_{\bm{\theta}}$ to be a linear function $f_{\bm{\theta}}(\myvec{x})=\myvec{\theta}^T\myvec{x}$, the above MMSE estimates are further reduced to

\begin{equation}\label{eq:mmse_stg_linear}
\begin{aligned}
    E[Y|X=\myvec{x},S'=\myvec{s'}] & = f_{\bm{\theta}^{*}}(\myvec{x} \odot \myvec{s'}) \\
    & = \myvec{\theta}^{*T}(\myvec{x}\odot \myvec{s'}) \\
    & = (\myvec{x} \odot \myvec{\theta}^{*})^{T}\myvec{s'}
\end{aligned}
\end{equation}
\begin{equation}\label{eq:mmse_cstg_linear}\begin{aligned}
    E[Y|X=\myvec{x},Z=\myvec{z},S'(\myvec{z})=s'(\myvec{z})] &= f_{\bm{\theta}^{*}}(\myvec{x}\odot {s'(\myvec{z})}) \\
    &= \myvec{\theta}^{*T}(\myvec{x}\odot {s'(\myvec{z})}) \\
    &= (\myvec{x} \odot \myvec{\theta}^{*})^{T}s'(\myvec{z})
    \end{aligned}
\end{equation}
Step 3: We now average the estimates given in Eq.~\eqref{eq:mmse_stg_linear} and Eq.~\eqref{eq:mmse_cstg_linear} across $S'$ and $S' \vert Z=\myvec{z}$ respectively.
\begin{equation}\label{eq:mmse_stg_linear_pi}
    E[Y|X=\myvec{x}] = E_{S'}(\myvec{x} \odot \myvec{\theta}^{*})^{T}\myvec{s'} = (\myvec{x} \odot \myvec{\theta}^{*})^{T}\myvec{\pi}^{*}_\textrm{stg}
\end{equation}
\begin{equation}\label{eq:mmse_cstg_linear_pi}
\begin{aligned}
    E[Y|X=\myvec{x},Z=\myvec{z}] &= E_{S'(\myvec{z})}(\myvec{x} \odot \myvec{\theta}^{*})^{T}s'(\myvec{z}) \\ &= (\myvec{x} \odot \myvec{\theta}^{*})^{T}\pi^{*}(\myvec{z})
\end{aligned}
\end{equation}
Step 4: We now project the MMSE estimate of c-STG onto the space of contextual-feature selection vectors that remain constant across contextual variables. This is given by
\begin{equation}\label{eq:mmse_cstg_linear_projection}
\begin{aligned}
    E_{Z}E[Y|X=\myvec{x},Z=\myvec{z}] &= E_{Z}(\myvec{x} \odot \myvec{\theta}^{*})^{T}\pi(\myvec{z}) \\ &= (\myvec{x} \odot \myvec{\theta}^{*})^{T} E_{Z}\pi^{*}(\myvec{z})
\end{aligned}
\end{equation}

Step 5: This solution should be equivalent to solving the STG problem as it restricts the stochastic gates to be constant over $z$. Comparing the two equations Eq.~\eqref{eq:mmse_stg_linear_pi} and Eq.~\eqref{eq:mmse_cstg_linear_projection}, we get $\myvec{\pi}^*_\textrm{stg} = E_{Z} [\pi^*(\myvec{z})]$.

\stgrelationmu*
\textit{Proof: }
 This follows from the proof of Theorem 3.3 by replacing Eq.~\eqref{eq:mmse_stg_linear_pi} and Eq.~\eqref{eq:mmse_cstg_linear_pi} with the following two equations respectively:
\begin{equation}\label{eq:mmse_stg_linear_mu}
    E[Y|X=\myvec{x}] = E_{S'}(\myvec{x} \odot \myvec{\theta}^{*})^{T}\myvec{s'} = (\myvec{x} \odot \myvec{\theta}^{*})^{T}\myvec{\mu}^{*}_\textrm{stg}
\end{equation}
\begin{equation}\label{eq:mmse_cstg_linear_mu}
\begin{aligned}
    E[Y|X=\myvec{x},Z=\myvec{z}] &= E_{S'(\myvec{z})}(\myvec{x} \odot \myvec{\theta}^{*})^{T}s'(\myvec{z}) \\ &= (\myvec{x} \odot \myvec{\theta}^{*})^{T}\mu^{*}(\myvec{z})
\end{aligned}
\end{equation}

\lowriskwcSTG*
\textit{Proof: } This follows similar to the proofs of the theorem~\ref{thm:low_risk_cSTG}. The feasible solution space of c-STG is contained in the feasible solution space of weighted c-STG as the weights of the weighted c-STG can be any function of contextual variables $\myvec{z}$, which includes a constant function, which is the case in c-STG. Therefore, the optimal risk attained by weighted c-STG must be less than or equal to the optimal risk value attained by c-STG.

\section{Additional Simulated Datasets}
\subsection{Moving XOR}
Here, we validate that the c-STG method can identify context-specific informative features while learning complex non-linear prediction functions.  
Following synthetic examples in \cite{yamada2020feature,yang2022locally}, we design a moving XOR dataset as 

\begin{align}
    \textbf{XOR1:} \qquad \textnormal{y}(\myvec{x},\myvec{z}) &=
    \begin{cases}
        {x}_1 \times {x}_2,& \text{if } \myvec{z} = 0,\\ 
       {x}_2 \times {x}_3 ,& \text{if } \myvec{z} = 1,\\
        {x}_3 \times {x}_4 ,& \text{if } \myvec{z} = 2,\\
    \end{cases} . \label{eq:exp6}
\end{align}

\begin{figure}[t]
    \centering
   \includegraphics[width=0.8\linewidth]{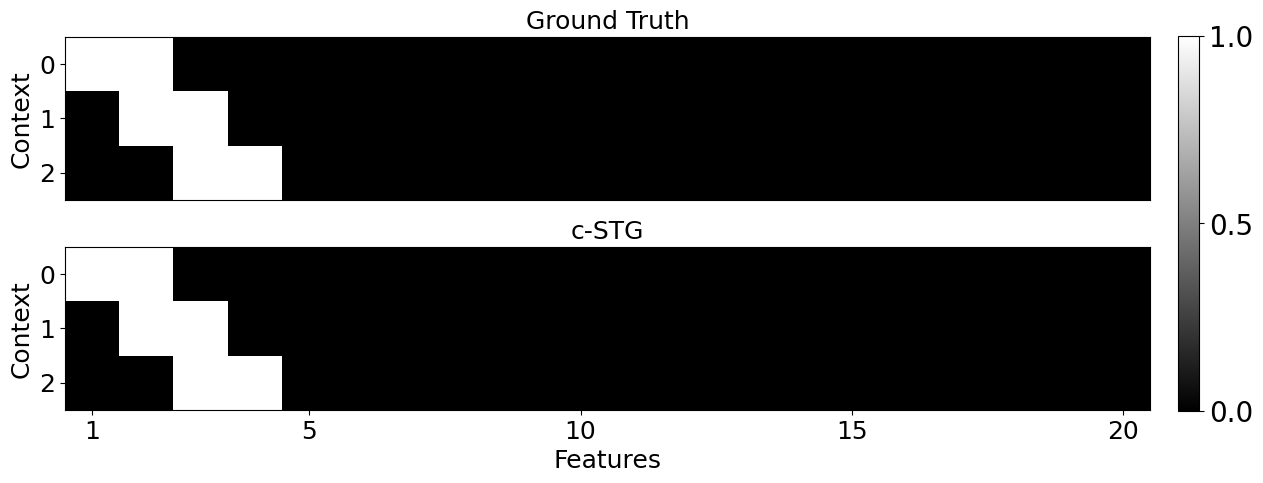}
    \caption{XOR1. The ground truth (top) feature importance as a function of $z$.
     Features identified by c-STG (bottom).}
    \label{fig:xor}
\end{figure}

We generate a data matrix ${X}$ with $1500$ samples and $20$ features, where each entry is sampled from a fair Bernoulli distribution ($P(x_{ij} = 1) = P(x_{ij} = -1) = 0.5$). 
The contextual variable $\myvec{z}$ is sampled uniformly from $\{0,1,2\}$ and
the response variable $\textnormal{y}$ for different samples will have different subsets of significant features.
In Fig.~\ref{fig:xor}, we demonstrate that c-STG correctly recovers the dependence of $y$ on the features as a function of $\myvec{z}$. 

\begin{figure}[t]
    \centering
   \includegraphics[width=\linewidth]{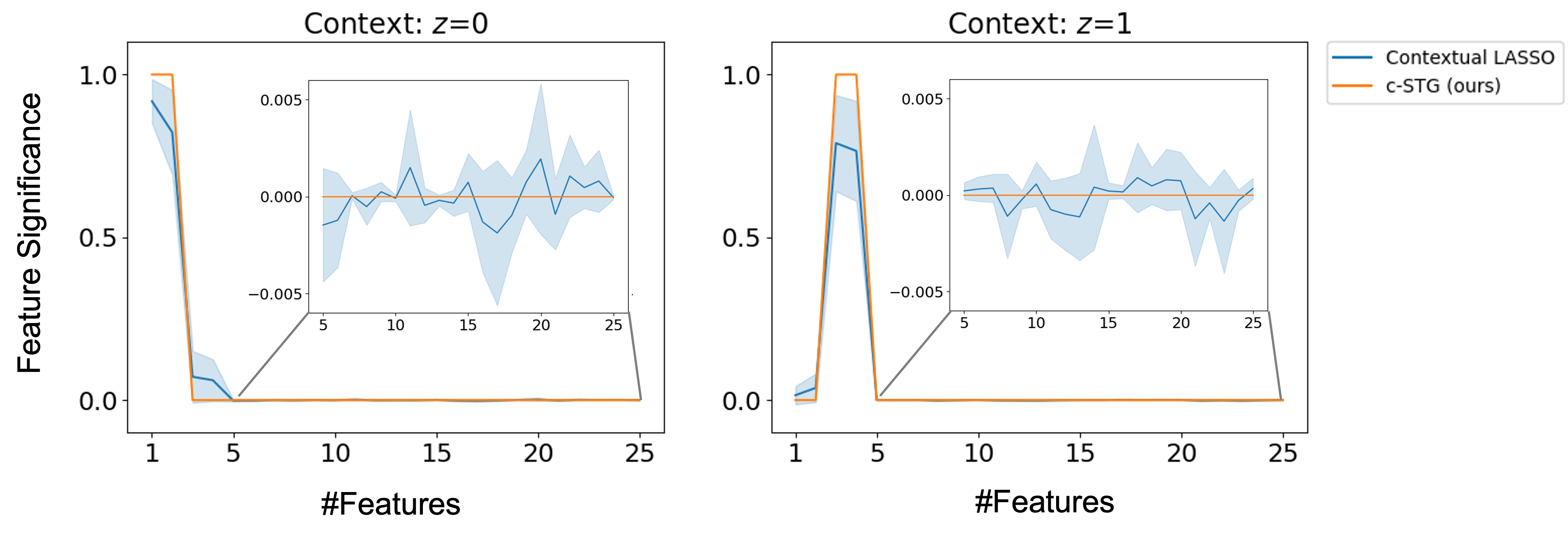}
    \caption{XOR3. Comparative feature significance between Contextual LASSO (blue) and c-STG (orange) for contexts $z=0$ (left) and $z=1$ (right). c-STG precisely isolates features 1 and 2 as significant for $z=0$ and features 3 and 4 for $z=1$, assigning no significance to other features. In contrast, Contextual LASSO indicates non-zero significance to all features, though it correctly identifies higher significance for features 1 and 2 for $z=0$ and similarly for $z=1$. This disparity showcases the shrinkage challenge of Contextual LASSO, which c-STG overcomes, ensuring accurate feature sparsity. }

    \label{fig:xor3}
\end{figure}

Additionally, to illustrate c-STG's superiority over contextual LASSO, a context-specific feature selection technique using $\ell_1$ regularization, we create a third synthetic example (XOR3), designed for linear prediction models where contextual LASSO is applicable. This example aims to showcase c-STG's advantage, particularly its resistance to the shrinkage problem that plagues contextual LASSO.
We construct a dataset ${X}$ with $1000$ samples and $25$ features, drawing each entry from a normal distribution. The contextual variable $\myvec{z}$ is uniformly sampled from ${0,1}$. The prediction variable $\textnormal{y}$ is defined as follows, where different subsets of features become significant depending on $\myvec{z}$'s value:
\begin{align*}
    \textbf{XOR3:} \qquad \textnormal{y}(\myvec{x},\myvec{z}) &=
    \begin{cases}
       {x}_1 + {x}_2 + \eta ,& \text{if } \myvec{z} = 0,\\
        {x}_3 + {x}_4 + \eta,& \text{if } \myvec{z} = 1,\\
    \end{cases} . 
\end{align*} \label{eq:exp7}
with $\eta \sim \mathcal{N}(0,0.25I)$. In Fig.~\ref{fig:xor3}, c-STG accurately identifies the significant features for each context, demonstrating its effectiveness compared to the contextual LASSO, which is hindered by coefficient shrinkage.

\begin{figure}[t]
    \centering
   \includegraphics[width=0.8\linewidth]{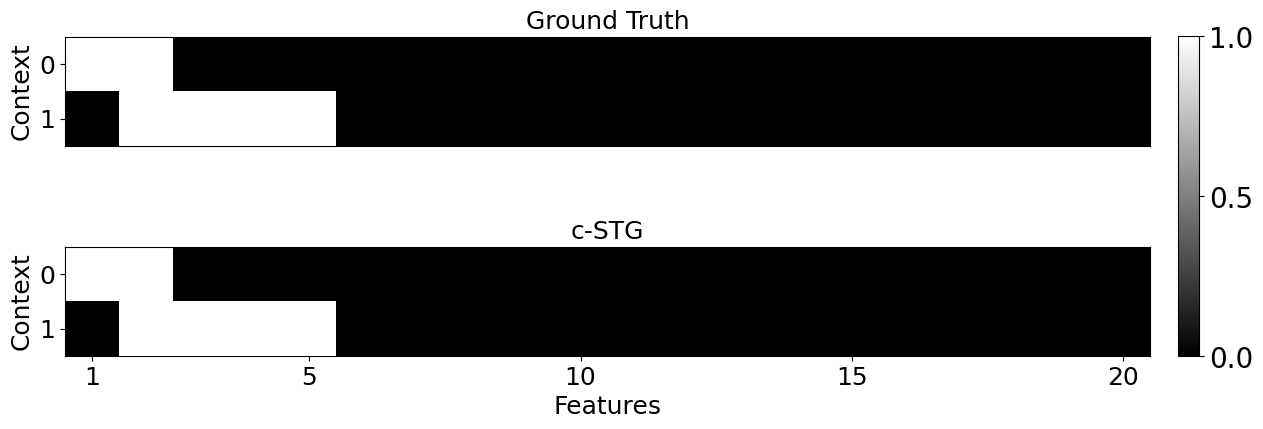}
    \caption{XOR4. The ground truth (top) feature importance as a function of $z$.
     Features identified by c-STG (bottom).}
    \label{fig:xor4}
\end{figure}

Furthermore, to show that c-STG can identify different important features for each context, we create another synthetic example named XOR4. Similar to XOR3, 
we construct a dataset ${X}$ with $1000$ samples and $25$ features, drawing each entry from a normal distribution. The contextual variable $\myvec{z}$ is uniformly sampled from ${0,1}$. The prediction variable $\textnormal{y}$ is defined as follows, where different number of features become significant depending on $\myvec{z}$'s value: 
\begin{align*}
    \textbf{XOR4:} \qquad \textnormal{y}(\myvec{x},\myvec{z}) &=
    \begin{cases}
       {x}_1 + {x}_2 + \eta ,& \text{if } \myvec{z} = 0,\\
        {x}_3 + {x}_4 + {x}_5 + {x}_6 + \eta,& \text{if } \myvec{z} = 1,\\
    \end{cases} . 
\end{align*} \label{eq:exp8}

with $\eta \sim \mathcal{N}(0,0.25I)$. In Fig.~\ref{fig:xor4}, we demonstrate that c-STG accurately recovers the dependence of $y$ on the features as a function of $\myvec{z}$. 

\section{Implementation Details} \label{sec:hyperparameters}
In selecting model architecture, we aim to maintain simplicity, interpretability, and linearity whenever feasible. We retained the same hypernetwork and prediction model architecture across various methods when possible. For the hypernetwork, we conducted experiments with one and two hidden layers, while for the prediction networks, we explored options ranging from zero (linear) to two hidden layers. The final model architectures for each of the datasets are mentioned below in their respective paragraphs. 
To determine the best hyperparameters, namely the learning rate ($\eta$) and regularization coefficient ($\lambda$), we performed a grid search over the following values: $\eta \in \{1e^{-1},5e^{-2},1e^{-2},5e^{-3},1e^{-3},5e^{-4},1e^{-4}\}$ and $\lambda \in \{1,5e^{-1},1e^{-1},5e^{-2},1e^{-2},5e^{-3},1e^{-3}\}$. The same set of values was used for the grid search across all the datasets. The selection of model parameters/hyperparameters was based on preventing issues like underfitting and overfitting and ensuring optimal 5-fold cross-validated performance.

\paragraph{XOR1:}  
The hypernetwork $\widetilde{h}_\phi$ has two fully connected layers, with 100 and 10 neurons, respectively. We employed ReLU as the activation function after the first layer and Sigmoid activation after the last layer. For the prediction model, we used two fully connected layers, with 10 and 10 neurons, respectively, and their corresponding nonlinearities, ReLU and sigmoid. 

\paragraph{XOR2:}  
Both the hypernetwork $\widetilde{h}_\phi$ and the prediction model have no hidden layers with inputs directly projecting to outputs. However, the hypernetwork has a Sigmoid activation, and the prediction model has a ReLU activation function in their output layers.

\paragraph{XOR3:}  
For the XOR3 dataset, the architecture of hypernetwork and prediction model is same as that of XOR2 except that the prediction model has no activation function. 

\paragraph{XOR4:}  
For the XOR4 dataset, the architecture of hypernetwork and prediction model is same as that of XOR3. 

\paragraph{MNIST:}
For the MNIST dataset, the configuration of the hypernetwork $\widetilde{h}_\phi$ differs, possessing two layers with 64 and 128 neurons. ReLU and Sigmoid activation functions are used for these two layers, respectively. The prediction model consists of layers with 128 and 64 neurons, coupled with the ReLU followed by Sigmoid activations.

\paragraph{Housing: } We employed a hypernetwork architecture similar to the one used in the MNIST example. However, we used a linear model to predict house prices in this case.
We divided the data into 10 train, validation, and test splits and conducted a grid search on $\eta$ and $\lambda$ for each split. We then selected the hyperparameters with the best validation performance. 

\paragraph{Heart disease}
The full list of features includes:
chest pain type (cp), resting blood pressure (trestbps), serum cholesterol levels (chol), fasting blood sugar (fbs), resting electrocardiographic results (restecg), maximum heart rate achieved (thalach), exercise-induced angina (exang), ST depression induced by exercise (oldpeak), the slope of the peak exercise ST segment (slope), the number of major vessels colored by fluoroscopy (ca), and thalassemia (thal). Among these features, cp, restecg, and slope are categorical and, hence, are converted to one hot encoding vector for feeding them into the prediction model. 
The hypernetwork and the prediction model have one hidden layer with 1000 neurons and 10 neurons, respectively, with sigmoid activation on their last layer. 
We conducted a grid search on $\eta$ and $\lambda$ from the range of values mentioned in the MNIST example.

\paragraph{Neuroscience: } For the first data set (context is time), a weighted c-stg model was trained using 5-fold cross-validation with separate sets of trials for training, parameter tuning, and testing. We conducted a grid search on different learning rates and regularization parameters ranging from 0.0005 to 0.1 and for the number of hidden units of the hypernetwork ranging from 10 to 1000 units.
Overall, for the hypernetwork, we used a single hidden layer of 10 units, a learning rate of 0.0005, and a regularization weight of 0.05, where the prediction network consisted of six hidden layers of 500, 300, 100, 50, and 2 units. 
For the second data set (flavor is context), a c-stg model was trained using 5-fold cross-validation with the same mechanism for parameter tuning. Overall, for the hypernetwork, we used a single layer with 1000 units, a learning rate of 0.05, and a regularization weight of 0.001, where the prediction network consisted of six hidden layers of 400, 300, 100, 50, and 2 units.

\section{Additional Experiments}

\begin{figure}[t]
    \centering
    \includegraphics[width=0.5\linewidth]{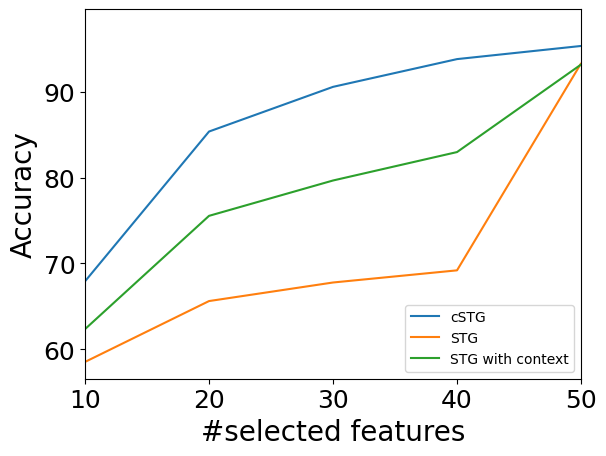}
    \caption{Comparing the performance variations of cSTG, STG, and STG with context with respect to different sparsity levels on the MNIST dataset.}
    \label{fig:sparsity}
\end{figure}

\paragraph{Sparsity-Driven Performance Analysis:}
We evaluate the performance of c-STG as a function of the number of selected features in the model, i.e., sparsity. Our analysis uses the MNIST example, as described in section 4. To perform this analysis, we varied the regularization coefficient, $\lambda$, in a way that controlled the sparsity level within the range of 10 to 50. It is worth noting that for STG with context, we only counted the explanatory features corresponding to pixels and not the eight additional contextual features (one-hot encoding) representing the angle.
Figure~\ref{fig:sparsity} illustrates the performance of three models, c-STG, STG, and STG, with context as the sparsity level increases. This demonstrates that c-STG has superior performance compared to both STG with context and STG alone, indicating that c-STG, with its contextual information and sparse representation, outperforms the other models for varying numbers of selected features.

\begin{table*}[t]
\centering
 \caption{Comparison of feature selection techniques with a fixed number of significant features across all the models. The number of significant features selected for each dataset is reported in the second row. \textbf{Bold} indicates best performance and \underline{underline} indicates second-best.}
\begin{tabular}{c|c||c||c||c||c|}
\cline{2-6} & \multicolumn{1}{c||}{\textbf{XOR1}}
 & \multicolumn{1}{c||}{\textbf{XOR2}} & \multicolumn{1}{c||}{\textbf{MNIST}}& \multicolumn{1}{c||}{\textbf{Heart disease}}  & \multicolumn{1}{c|}{\textbf{Housing}}                        \\ \cline{2-6}
 & 3 & 2   & 218  & 7 & 5   \\ \cline{2-6}
 & Accuracy ($\%$ ) & $R^2$ score   & Accuracy ($\%$ )  & Accuracy ($\%$) & $R^2$ score   \\ \hline
\multicolumn{1}{|c|}{LASSO}          & 50.07 (1.43) & 0.1760 (0.0156) & 73.70 (0.48)   &  73.70 (0.48)    & 0.1979 (0.0016)
                             \\ 
\multicolumn{1}{|c|}{with context}   & 50.02 (1.09)  & 0.1761 (0.0156)
 & 73.79 (0.48)
   &  82.00 (3.48)
   & 0.2251(0.0015) \\ \hline

\multicolumn{1}{|c|}{STG }             & 65.73 (1.46)  & 0.1838 (0.0248)
 & 91.47 (0.67)   & 84.18 (2.92)  & 0.2136 (0.0026)                        \\ 
 
\multicolumn{1}{|c|}{with context}     & 65.77 (0.87) & 0.1873 (0.0320) & 98.34 (0.07)    & 83.84 (1.31)  & 0.2231 (0.0022)                      \\ \hline

\multicolumn{1}{|c|}{Contextual LASSO }   & 50.53 (0.60) & 0.7311 (0.0090) & 97.08 (0.06) & 82.85 (4.83)
 & \ul{0.4535} (0.0034)
                       
    \\ \hline
\multicolumn{1}{|c|}{AFS }                 & \ul{70.83} (19.06) & 0.7572 (0.0187) & 93.91 (0.99)  & 83.39 (1.98)
  & 0.3482 (0.0224)
              \\ \hline

\multicolumn{1}{|c|}{INVASE }              & 50.07 (0.70) & 0.3547 (0.0118) & 65.27 (4.69) & 84.18 (1.72) & 0.2502 (0.0060)
 \\ \hline

\multicolumn{1}{|c|}{c-STG (Ours) }        & \textbf{100} (0)  &  \ul{0.8739} (0.0107) & \ul{98.65} (0.04) &  \ul{86.89} (3.18) & 0.3970 (0.0074)           \\ 
\hline

\multicolumn{1}{|c|}{Weighted c-STG (Ours) }   & \textbf{100} (0)    & \textbf{0.9956} (0.0008) & \textbf{98.69} (0.06)        & \textbf {89.23} (1.72)          & {\textbf {0.5308}} (0.0052)           \\ 
\hline

\end{tabular}
\label{tab:results3}
\end{table*}

\begin{figure}[t]
    \centering
    \includegraphics[width=0.5\linewidth]{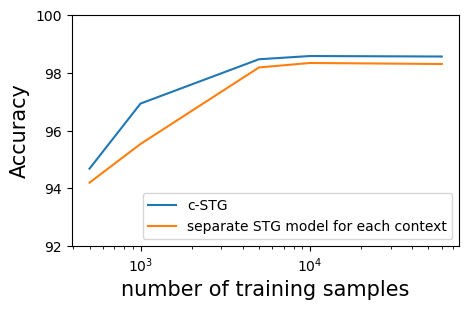}
    \caption{Comparison between training separate models for different values of categorical context variables and a single c-STG model on the modified MNIST dataset in section 4, as the number of training samples varies.}
    \label{fig:num_samples_accuracy}
\end{figure}

\paragraph{Comparison of Discrete Contextual Model Training:}
When focusing on deriving insights from specific contextual features, training separate models for each specific context is a viable strategy. However, it is crucial to understand that this strategy is only applicable when the contextual variables are categorical and not continuous. Although straightforward, this method brings its own challenges, particularly when the dataset at hand is limited in size. In such cases, each model only gets to train on a subset of data corresponding to its designated context. We shed light on the consequences of this approach using the modified MNIST dataset, as elaborated in Section 4. Our exploration involved comparing two methodologies: 1) individually training eight models for each context using STG and 2) employing our c-STG to create a unified model for all contexts. To assess the efficiency of these approaches, we varied the number of training samples and captured the results in Fig.~\ref{fig:num_samples_accuracy}.

Comparatively, our c-STG model offers a compelling advantage. Instead of segregating data by context, it utilizes the entirety of the dataset, benefiting from the diverse range of contexts. A direct contrast in terms of model parameters reveals that while training separate models for each discrete context demands 876,688 parameters, the c-STG model operates with a lean set of 218,834 parameters. The reduced parameter count in c-STG does not compromise performance; in fact, it underscores c-STG's prowess in delivering enhanced outcomes with fewer parameters.

\paragraph{Feature Selection Across Models:}
We present a comparative analysis detailing the number of features chosen by each method whose prediction performance is reported in Table~\ref{tab:results}. A notable observation is that c-STG has fewer selected features and ranks either as the best or second-best in performance evaluation, as demonstrated in Table~\ref{tab:results}. The prediction model (with context) and CEN are excluded, as no sparsity constraint is imposed on the explanatory features to perform the prediction task.

\section{Compute details}
We trained all networks using CUDA-accelerated PyTorch implementations on a NVIDIA Quadro RTX8000 GPU.

\end{document}